\documentclass[conference]{IEEEtran}
\ifCLASSINFOpdf
\else
\fi
\let\labelindent\relax
\usepackage{tikz}
\usepackage{enumitem}
\usepackage{graphicx}
\usepackage{bm}
\usepackage{multirow}
\usepackage{array}
\usepackage{booktabs}
\usepackage{makecell}
\usepackage{hyperref}
\usepackage{soul}
\usepackage{subcaption,graphicx}
\usepackage{amsmath}
\usepackage[misc,geometry]{ifsym}

\usepackage{hyperref}

\newcommand{\mct}[1]{({\em #1})}

\newcommand{\SLJ}[1]{\textcolor{black}{#1}}

\newcommand{\paragraphbe}[1]{\smallskip\noindent{\bf {#1}.}~}

\begin{document}
\title{Improving the Robustness of Transformer-based Large Language Models with Dynamic Attention}

%\author{\IEEEauthorblockN{Lujia Shen, Yuwen Pu, Shouling Ji}
%\IEEEauthorblockA{Zhejiang University\\
%shen.lujia@zju.edu.cn}
%\and
%\IEEEauthorblockN{Changjiang Li}
%\IEEEauthorblockA{Penn State\\
%yw.pu@zju.edu.cn}
%\and
%\IEEEauthorblockN{Zhejiang University}
%\IEEEauthorblockA{Starfleet Academy\\
%someemail@somedomain.com}}



\author{\IEEEauthorblockN{Lujia Shen\IEEEauthorrefmark{1},
Yuwen Pu\IEEEauthorrefmark{1}\textsuperscript{, \Letter}\thanks{\Letter \ Yuwen Pu is the corresponding author.},
Shouling Ji\IEEEauthorrefmark{1},
Changjiang Li\IEEEauthorrefmark{2},
Xuhong Zhang\IEEEauthorrefmark{1}, 
Chunpeng Ge\IEEEauthorrefmark{3} and
Ting Wang\IEEEauthorrefmark{2}}
\IEEEauthorblockA{\IEEEauthorrefmark{1}Zhejiang University, \IEEEauthorrefmark{2}Penn State, \IEEEauthorrefmark{3}Shandong University}
\IEEEauthorblockA{Emails: \{shen.lujia, yw.pu, sji, zhangxuhong\}@zju.edu.cn, \{meet.cjli, inbox.ting\}@gmail.com, gechunpeng2022@126.com}
}


\IEEEoverridecommandlockouts
\makeatletter\def\@IEEEpubidpullup{6.5\baselineskip}\makeatother
\IEEEpubid{\parbox{\columnwidth}{
    Network and Distributed System Security (NDSS) Symposium 2024\\
    26 February - 1 March 2024, San Diego, CA, USA\\
    ISBN 1-891562-93-2\\
    https://dx.doi.org/10.14722/ndss.2024.24115\\
    www.ndss-symposium.org
}
\hspace{\columnsep}\makebox[\columnwidth]{}}

\maketitle

%\footnote{\blank}{{\Letter}Yuwen Pu is the corresponding author}

\begin{abstract}
%\boldmath

Transformer-based models, such as BERT and GPT, have been widely adopted in natural language processing (NLP) due to their exceptional performance. However, recent studies show their vulnerability to textual adversarial attacks where the model's output can be misled by intentionally manipulating the text inputs. Despite various methods that have been proposed to enhance the model's robustness and mitigate this vulnerability, many require heavy consumption resources (e.g., adversarial training) or only provide limited protection (e.g., defensive dropout). 

In this paper, we propose a novel method called dynamic attention, tailored for the transformer architecture, to enhance the inherent robustness of the model itself against various adversarial attacks. Our method requires no downstream task knowledge and does not incur additional costs. The proposed dynamic attention consists of two modules: \mct{i} {\em attention rectification}, which masks or weakens the attention value of the chosen tokens, and \mct{ii} {\em dynamic modeling}, which dynamically builds the set of candidate tokens. 
Extensive experiments demonstrate that dynamic attention significantly mitigates the impact of adversarial attacks, improving up to 33\% compared to previous methods against widely used adversarial attacks. The model-level design of dynamic attention enables it to be easily combined with other defense methods (e.g., adversarial training) to further enhance the model's robustness. Furthermore, we demonstrate that dynamic attention preserves the state-of-the-art robustness space of the original model compared to other dynamic modeling methods.
\end{abstract}


% \fi
%%\IEEEpeerreviewmaketitle

\section{Introduction}
Transformer architectures have gained widespread research interest recently, especially in natural language processing (NLP) fields \cite{LIN2022111}. Compared to traditional NLP models like RNNs, transformers leverage attention mechanism to effectively capture long-range dependencies in sequences and better handle the variable-length input \cite{zhuang2022long, zhou2021informer}, making it achieve outstanding performance on a wide range of NLP tasks, including text generation \cite{scao2022bloom}, text classification \cite{devlin2019}, and question answering \cite{zhao2020condition}. Moreover, many Internet giants have trained large-scale transformer-based language models (i.e., {\em foundation models}), such as GPT-3 from OpenAI and T5 from Google, which can be adapted to a variety of downstream tasks through different tuning strategies (e.g., fine-tuning and prompt-tuning) \cite{liu2023pre}. Despite their generalizability \cite{brutzkus2019larger}, these foundation models remain susceptible to malicious input perturbations \cite{li2019textbugger, jin2020bert}, which can lead to undesirable outcomes in target NLP models, such as the generation of abusive content in machine translation or the misclassification of toxic comments as benign in toxicity classification \cite{zeng2021empirical}.

To counteract the threat posed by adversarial attacks in NLP, researchers have proposed various defense methods, broadly classified into empirical defenses and certified robust approaches \cite{li2022sok}. Adversarial training, one of the most effective and widely used empirical defenses, involves training the model to classify clean and adversarial examples correctly. 
However, as it requires generating adversarial examples during the training, it requires downstream task knowledge \cite{suresh2021adversarial, hsu2022adversarial, liu2020adversarial}, as it can be computationally expensive \cite{yoo2021towards}, especially for large foundation models. 

On the other hand, certifiably robust approaches include robustness verification to provide the lower bound of robust space against any attacks and the corresponding robust training method to enlarge the model's robustness space \cite{huang2019achieving, 9551763}.
However, certified robust training degrades the model's performance on the original tasks \cite{krishnan2020lipschitz, yang2020closer} and is hard to be generalized to all types of attacks \cite{cohen2019certified} due to the discrete input space in the context of NLP.
Moreover, for large foundation models, certified robust training needs an extremely long running time and easily outputs trivial bounds \cite{li2022sok}. 
These time-consuming and resource-intensive algorithms are difficult to provide a feasible robustness improvement scheme for practical large language models. 

Another emerging research topic on defending against adversarial attacks is the dynamic neural networks \cite{goodfellow2019research} that change each time they run. 
Previous dynamic models focus on adding randomness to hidden features like SAP \cite{s.2018stochastic} and defensive dropout \cite{wang2018defensive}. 
However, these methods only consider feature-level operations, which may result in the loss of information.
Additionally, there is a lack of dynamic modeling methods focusing on the feature extraction process, especially in the attention mechanism. 

To address the above-mentioned shortcomings, in this paper, we propose dynamic attention, tailored for transformer-based models, to mitigate the effect of adversarial attacks, which can be adapted to different transformer architectures and different downstream tasks without any downstream task knowledge, and can be combined with other defense modules for further improvement.

We first explore the impact of adversarial examples on the attention mechanism. 
Our empirical findings reveal that in adversarial examples, task-irrelevant tokens are often assigned higher attention values, which differ from the attention values assigned to relevant tokens in clean texts.
%However, tokens with higher attention values in clean samples are more task-related. 
In addition, our results demonstrate that by substituting the attention of an adversarial text with the attention of its corresponding clean text, the model can correctly classify the adversarial text. 
This suggests that assigning appropriate attention values to each input token is crucial for achieving correct classification.
These two observations indicate that adversarial examples can mislead the attention mechanism, leading to model's misbehavior. 
These results stimulate the idea of mitigating adversarial attacks by rectifying the attention mechanism, which has not been investigated by prior works. 
Thus, in this paper, we propose attention rectification to mask or weaken the attention values of the key tokens with greater attention.
To determine the set of tokens with greater attention, we accumulate the attention map of all heads in each layer and rank each key token's total attention accordingly.
Furthermore, based on the difference in adversarial attacks between text classification and text generation, we propose two different token set selection rules to optimize attention.

Next, previous works like \cite{tian2021detecting} and our empirical experiments reveal that adversarial examples are inherently unstable; that is, the model is sensitive to adversarial examples. 
Moreover, a fixed number of weakened or masked tokens cannot reduce the attack success rate of adversarial attacks, as only the number of attack queries increases.
Thus, we introduce dynamic modeling to attention rectification and propose dynamic attention. 
Specifically, we let the rectified attention have different numbers of token modifications in each transformer layer and in each time they run.
Dynamic attention is also compatible with dropout, where the fusion model composed of the two can further improve the model's resistance to attack.

Finally, we set up and experiment with three different attack scenarios based on the attacker's capabilities and knowledge of the target model. 
We also conduct experiments using three representative attacks on three different tuning methods.  
Experimental results show that dynamic attention is effective in all three text classification datasets and two text generation datasets. 
In addition, dynamic attention can also mitigate the attacks on the shifted dataset and attenuate the effect of backdoor attacks.
\SLJ{Further experiments like stability analysis in Sec. \ref{sec:stable} and statistical robustness analysis of dynamic attention in Sec. \ref{sec:analysis} show that the dynamic attention model produces more stable output than the dropout and fusion models and can preserve the robustness of the original model.}

\textbf{Contributions.} In summary, we make the following contributions in this paper.
\begin{enumerate}[leftmargin=*]
\item We propose a dynamic attention mechanism, which, to our best knowledge, is the first dynamic modeling method tailored for transformer-based models. It modifies the crucial attention mechanism to help weaken the influence of adversarial examples on the output, thereby improving the model's robustness without prior knowledge. \SLJ{It is also compatible to different transformer architectures, including encoder-only model, decoder-only model and encoder-decoder model.} It serves as a supplementary to existing robustness-enhancement methods instead of an alternative. 
%\item The dynamic attention mechanism requires no knowledge about the downstream datasets and is applicable to different NLP tasks which is critical to large foundation models.   \bluetext{this is an advantage, not contribution}
\item We evaluate the effectiveness of the dynamic attention mechanism on different tasks (text classification, text generation) under multiple attacks (TextBugger \cite{li2019textbugger}, TextFooler.\cite{jin2020bert}, PWWS \cite{ren2019generating}). Evaluation results show that the dynamic attention mechanism is effective in different settings (three tuning methods, three attack threat models, three classification, and two generation tasks), and combining dynamic attention with dropout can further improve the model's robustness up to 33\% than the defensive dropout model. It can also work concurrently with other modules (e.g., information bottleneck and adversarial training) to further enhance the model's robustness. 
% \bluetext{how much?  xxx\%.}
\item At the same time, our experiments demonstrate that the dynamic attention mechanism is effective in defending adaptive attacks and can preserve 98\% of the model's robustness and output more stable predictions than the dropout and fusion models. 
\end{enumerate}

\section{Related Work}

\subsection{Transformer-based Large Language Models}\label{sec:lm}

The transformer model architecture has become a dominant player in language modeling. The introduction of the transformer models in Vaswani et al. \cite{vaswani2017attention} marked a significant leap in the field of NLP with their outstanding performance. Large language models built on the transformer architecture, such as GPT \cite{brown2020language} and BERT \cite{devlin2019}, have outshone prior state-of-the-art models \cite{gillioz2020overview} thanks to their effective capturing of inter-word relationships using the scaled dot-product attention mechanism. \SLJ{Based on the model architecture, there are three types of transformer models: encoder-only model (BERT), decoder-only model (GPT), and encoder-decoder model (T5).}

\SLJ{These large language models are typically pre-trained on a large unlabeled corpus and then adapted to downstream tasks using three strategies: full parameter fine-tuning (fine-tuning) and parameter efficient fine-tuning techniques such as prefix-tuning and prompt-tuning. }
\mct{i} \SLJ{{\em Full parameter fine-tuning (Fine-tuning)} \cite{howard2018universal}} is the most widely used method, which updates the entire pre-trained model for a new task. However, it is computationally expensive as it requires retuning all parameters for each new task, even if only a small portion is relevant. 
\mct{ii} {\em Prompt-tuning} \cite{lester2021power} prepends learnable token embeddings (called soft prompts) before the input to condition frozen language models to perform downstream tasks, leading to improved performance and reduced computational cost. \mct{iii} {\em Prefix-tuning} \cite{li2021prefix} optimizes a set of task-specific vectors (called the prefix), one for each layer, while maintaining the language model parameters constant. 
Unlike prompt-tuning, which is only associated with the input layer, prefix-tuning optimizes the prefixes associated with each layer, providing more trainable parameters. 
%Prompt-tuning can be seen as a simpler version of prefix-tuning with fewer parameters. 
In our work, we evaluate the effectiveness of our method using all three tuning strategies.

%\bluetext{Do we need to introduce the attention mechanism? This paper is a variant of the attention mechanism.}

\subsection{Textual Adversarial Attacks}
A textual adversarial attack involves the deliberate manipulation of text inputs to mislead an NLP model into producing an incorrect or undesirable outcome (e.g., misclassification) while preserving the semantic meaning from a human perspective \cite{zhang2020adversarial}. Adversarial attacks in NLP can be categorized into three categories based on the method of transformation used in the attacks, which are character-level, word-level, and sentence-level attacks, according to recent study in this field \cite{wang2021towards}. Gao et al. \cite{gao2018black} introduced DeepWordBug, a character-level adversarial attack designed for the black-box scenario. The method calculates the significance of words in a text and generates imperceptible perturbations by altering the selected words through techniques such as swapping, flipping, inserting, and deleting. In a character-level attack, only characters of the chosen words will be perturbed. 

On the other hand, word-level attacks replace entire words to achieve the attack goal. Papernot et al. \cite{papernot2016crafting} generated adversarial examples in a white-box scenario using the fast gradient sign method \cite{goodfellow2014explaining}. However, the substitution words chosen using FGSM may contain grammatical errors. \SLJ{Recent works have focused on word-level attacks through synonym substitution, as elaborated below. Jin et al. \cite{jin2020bert} proposed TextFooler, a black-box adversarial attack aimed at fooling the text classification model. The method identifies crucial words in a text and replaces them with synonyms. Similarly, Ren et al. \cite{ren2019generating} developed PWWS, which also uses synonym substitution to generate adversarial examples.} The difference between TextFooler and PWWS lies in how synonyms are chosen. Li et al. \cite{li2019textbugger} proposed TextBugger, a multi-level attack that combines character-level and word-level attacks.

Sentence-level attacks usually involve inserting sentences with unchanged semantics and grammar into the text. This type of attack is rarely used in classification tasks and can be seen as a special case of word-level attacks \cite{wang2021towards}. 
\SLJ{Therefore, in our experiments, we only consider character-level attacks and word-level attacks.}

\subsection{Improving Model Robustness}

%\bluetext{Rewrite: empirical (adversarial, distillation, dynamic modeling) and certified. Refer to acl.}

A large amount of defense works on model robustness enhancement have been proposed to mitigate the threat posed by textual adversarial attacks, and they can be classified into two categories: empirical defenses, such as empirical adversarial training and dynamic modeling, and certified robust approaches, such as certified robust training.

\paragraphbe{Empirical Adversarial Training} 
Goodfellow et al. \cite{goodfellow2014explaining} first proposed adversarial training when they proposed the FGSM attack algorithm. 
Based on this attack algorithm, they added adversarial examples generated by FGSM to the training set for training.
Vanilla adversarial training in NLP augments the training text with adversarial texts perturbed in the input space \cite{yoo2021towards}. 
In contrast, non-vanilla adversarial training performs perturbations on non-input space, such as word embeddings.
Zhu et al. proposed FreeLB \cite{zhu2019freelb}, which adds adversarial perturbations to word embeddings and minimizes the resultant adversarial loss around input samples.
Liu et al. proposed ALUM \cite{liu2020adversarial} that perturbs the embedding space and is applicable to pre-training.
Zhang et al. proposed an information bottleneck (IB) layer \cite{zhang2022improving} to better capture the robust textual features of FreeLB.
%Balunovi{\'c} et al. proposed a certified adversarial training, CLOT \cite{balunovic2020adversarial}, to enlarge the model's certified robust space.
For large-scale pre-trained models, generating adversarial examples requires repeated queries of the model, which consumes many resources. 
\SLJ{Moreover, to obtain knowledge from adversarial examples, it is usually necessary to update all model parameters \cite{REN2020346}, which may reduce the model's generalization ability \cite{raghunathan*2019adversarial, ijcai2021p591, clarysse2023why}. }
Besides, the number of trainable parameters in the prompt learning scenario is very limited, and it is difficult to carry the information introduced by adversarial training. 
Finally, adversarial training relies on specific downstream tasks to generate adversarial examples.
However, pre-trained models are trained from unsupervised learning.
Hence, adversarial training has limitations in usage scenarios.

%\paragraphbe{Defensive Distillation}
%Papernot et al. \cite{papernot2016distillation} proposed a defensive distillation algorithm where they train the distillation model using training samples and their output probability distribution of the base model. 
%\SLJ{However, Soll et al. \cite{soll2019evaluating} demonstrated that defensive distillation does not help with improving the robustness of their neural networks against adversarial examples.}
%Additionally, defensive distillation requires training two models and relies on soft labels given by downstream classification tasks. 
%However, large foundation models are usually trained through unsupervised training, so defensive distillation largely limits the type of large models. 
%Moreover, defensive distillation cannot defend against adversarial attacks that use gradients after the softmax layer. 
%Compared with this method, this method is applicable to various attack algorithms without additional training overhead.

\paragraphbe{Dynamic Modeling}
%\jiang{first?}
The dynamic neural network is an emerging research topic in deep learning. 
Compared to static models, which have fixed computational graphs and parameters at the inference stage, dynamic models can adapt their structures or parameters to different inputs, leading to notable advantages in terms of accuracy, computational efficiency, adaptiveness, etc. \cite{han2021dynamic}.
Early works towards dynamic modeling in NLP usually focus on accelerating the inference efficiency by adjusting the model's computational path based on the input like skimming techniques \cite{yu2018fast} and early exit \cite{xin2021berxit} instead of providing robustness enhancement. 

Recent works discovered that dynamically changing the model parameters or structure can obfuscate the gradients, which can serve as a defense mechanism against adversarial attacks.  
Dhillon et al. proposed Stochastic Activation Pruning (SAP) \cite{s.2018stochastic}, which stochastically prunes a subset of the activations in each layer, preferentially retaining activations with larger magnitudes.
In defensive dropout \cite{wang2018defensive}, Wang et al. proposed to use dropout at test time to achieve strong defense effects.
While Athalye et al. \cite{athalye2018obfuscated} and Yue et al. \cite{yue_gradient_nodate} demonstrated that gradient obfuscation could be circumvented by approximating the derivatives, their focus was primarily on the computer vision (CV) domain, which predominantly deals with continuous input spaces.
In contrast, NLP models have discrete input space, rendering gradient reconstruction and correspondence to actual tokens challenging during deployment stages. 
In our experiments, we will compare our method with the defensive dropout.

\paragraphbe{Certified Robust Training}
Different robust training approaches are proposed to improve the model's robustness guarantees, typically related to or derived from the corresponding verification approach \cite{li2022sok}.
Shi et al. first introduced robustness verification to Transformers and improved \cite{Shi2020Robustness} and accelerated the certified robust training based on interval-bound propagation \cite{shi2021fast}. 
Wang et al. \cite{wang2021macrobert} used randomized smoothing to train BERT with maximized certified robust space.
However, deterministic verification using linear relaxation will lead to a trivial bound in large foundation models. 
Also, probabilistic verification has to construct a smoothed classifier, which is infeasible for large language models.

\section{Method}
%-------------------------------------------------------------------------------
%In this section, we first introduce the critical insight and then elaborate on the design of our defense modules. 
\subsection{Design Intuition}\label{sec:design_intuition}
%\bluetext{Rewrite}
We first conduct three empirical studies to illustrate the properties of adversarial texts and their implications for our method.
%, with detailed experimental settings provided in Appendix \ref{appendix:intuition1}, \ref{appendix:intuition2} and, \ref{appendix:intuition3}.

\begin{table}[t]
\centering
\caption{The prediction confidence difference between the attentive tokens of adversarial texts and their original texts.}
\scalebox{1.}{\begin{tabular}{ccccc}
\toprule
      Dataset             & Original & TextBugger & TextFooler & Average\\
                      \hline
Amazon & 0.1899 & 0.3618     & 0.3807   & 0.3713 \\
Twitter & 0.0059 & 0.5458     & 0.5152 & 0.5305\\
\bottomrule  
\end{tabular}}\label{tab:insights1}
\vspace{-5mm}
\end{table}

\paragraphbe{Tokens with high attention value in adversarial texts are different from those in their original texts}
Above all, to gain a comprehensive understanding of the characteristics and behaviors of adversarial texts in Transformer-based models, we analyze different performances of attention mechanism on clean and adversarial text.
One notable observation is that tokens with high attention values (attentive tokens) in the adversarial text exhibit notable differences compared to their counterparts in the original text. 
Specifically, the attentive tokens in the adversarial text are predominantly task-irrelevant, while the attentive tokens in the original text are primarily task-related.
To quantitatively demonstrate this result, we generate several adversarial texts using the Amazon dataset with a sentiment classification model for illustration.
We calculate the total attention received by each key token based on the attention maps and collect the model's attentive tokens, specifically the top five tokens with the highest attention values in each layer from the last six layers for both the adversarial texts and their original texts.
To demonstrate the semantic differences between attentive tokens from the clean text and those from the adversarial counterpart, we first concatenate these extracted tokens from the clean text to form a single, representative sentence. 
Then, we employ the classification model to predict the sentiment of this constructed sentence, and obtain the confidence. 
We also conduct the same operation to the attentive tokens extracted from the adversarial text. To assess the disparity in sentiment between the merged sentences from adversarial texts and those from their original texts, we computed the averaged difference in their prediction confidence, as shown in Table \ref{tab:insights1}.
%We provide more detailed experimental settings in Appendix \ref{appendix:intuition1}.
Additionally, we calculate the average confidence difference between merged sentences from clean texts and their 10\% random masked versions, indicated in the `Original' column, as a reference. A larger difference indicates a greater disparity in sentiment between the two merged sentences.

The results presented in Table \ref{tab:insights1} indicate that the difference between the merged sentences from adversarial texts and their original texts is more significant than between merged sentences from clean texts and their masked versions. 
For instance, in the Amazon dataset, there is a 0.3713 confidence difference between the attentive tokens in the adversarial text and their corresponding original text, which is significantly larger than the random masking. 

%We observe from Table \ref{tab:insights1} that the difference between merged sentences from adversarial texts and those from their original texts is larger than the difference between merged sentences from clean texts and their masked version. 
%For instance, in the Amazon dataset, there is a 0.37 confidence difference between the attentive tokens in the adversarial text and the corresponding original text. 
%If the merged sentence of attentive tokens from a clean text is predicted with 0.9 confidence, which contains sufficient semantics, that of adversarial text will be predicted with roughly 0.53 confidence, which contains barely no semantics. 

Similarly, we employ another toxic detection model to predict the toxicity confidence of the newly formed sentence by concatenating the attentive tokens in the Twitter dataset. The average difference in confidence between the adversarial and clean attentive tokens in the Twitter dataset is 0.5305, surpassing the difference in toxicity confidence between the attentive tokens from different original texts.
In conclusion, our results demonstrate that the attentive tokens in adversarial texts significantly differ from those in the original texts. 
This conclusion leads us to speculate that adversarial examples may influence the attention mechanism, causing the model to misbehave.

\paragraphbe{Replacing the attention of the adversarial text with the attention of its original text helps the model correctly classify the text}
%Since we have found that the model's attentive tokens of the adversarial text are quite different from those of their original text, we hypothesize that the adversarial examples mislead the attention mechanism and cause the model to misclassify them. 
To answer whether the adversarial texts take effect by misleading the attention mechanism, we consider replacing the attention of the adversarial text with the attention of its original text to investigate whether such attention replacement can help the model classify the text correctly.
Specifically, we keep the adversarial texts whose tokens can correspond one-to-one to the tokens of their original texts, where the attention maps of both can be aligned.
Then, we extract benign attention maps of the original text on all model layers. 
Next, the model's attention maps of the adversarial text are replaced by the extracted benign attention maps without calculating the attention through the dot-product of the query and key matrices.
Subsequently, the model outputs the prediction of the adversarial text with benign attention. 
Finally, we record the percentage of adversarial texts classified correctly in Table \ref{tab:insights2}.
Note that these adversarial texts all successfully attacked the model; that is, the prediction accuracy is 0\%.
%We defer more detailed experimental settings to Appendix \ref{appendix:intuition2}.

Table \ref{tab:insights2} shows that for the fine-tuned model, 88\% of adversarial texts, on average, are classified correctly into their original label by the target model. 
Similarly, 97\% of adversarial texts in the prompt-tuned model and 80\% of adversarial texts in the prefix-tuned model can be classified correctly with the replaced attention. 
That is, the model predicts most adversarial examples correctly when using the benign attention map. 
Therefore, we can conclude that the adversarial example misleads the attention mechanism in the transformer-based model and thus leads to the model's misbehavior. 
The above result induces us to modify the attention mechanism to force the model to pay less attention to those tokens with high attention values. 

\begin{table}[t]
\centering
\caption{The prediction accuracy of adversarial texts with attention replaced by their benign version.}
\scalebox{1.}{\begin{tabular}{cccc}
\toprule
Tuning Method & TextBugger & TextFooler & PWWS   \\
              \hline
Fine-tuning   & 86.96\%     & 90.62\%     & 87.27\% \\
Prefix-tuning & 82.61\%     & 80.65\%     & 75.81\% \\
Prompt-tuning & 94.11\%     & 95.65\%     & 100.0\% \\
\bottomrule   
\end{tabular}}
\label{tab:insights2}
\vspace{-5mm}
\end{table}

\paragraphbe{Most adversarial examples are inherently unstable} 
%\bluetext{conflict with results}
Previous studies \cite{jang2022strengthening, wu2021improving, 10.1007/978-3-030-37337-5_25} have shown that adversarial examples experience a trade-off between transferability and imperceptibility, i.e., the imperceptible adversarial examples generated from the surrogate model can hardly fool the target model. 
To show that adversarial texts also have low transferability, we first train two models from the same training dataset, generate adversarial texts from one model, and then transfer them to another.
We report the transfer rate of adversarial examples (i.e., the percentage of adversarial examples that the other model misclassifies) from the Amazon and Enron dataset in Table \ref{tab:insights3}. 
%Please refer to Appendix \ref{appendix:intuition3} for more detailed experimental settings.
Table \ref{tab:insights3} shows that TextFooler's adversarial transfer rate is only 41\% on Amazon. That is, 41\% of the adversarial examples can fool the model trained from the identical dataset. 
TextBugger's adversarial transfer rate is only 29\% on Twitter.
Therefore, adversarial texts are unstable and difficult to transfer well between different models even if they are trained from the same data.

The low transferability of the adversarial examples inspired us to use dynamic modeling, which can make the adversarial examples difficult to re-apply to the same model by changing the parameters of the model during each run \cite{goodfellow2019research, qin2021dynamic}.
Hence, we employ dynamic modeling in Transformer-based models to defend against adversarial transfer attacks.

\begin{table}[t]
\centering
\caption{The transferability rate of adversarial texts under models trained from the same data.}
\scalebox{1.}{\begin{tabular}{cccc}
\toprule
Dataset & TextBugger & TextFooler & PWWS    \\
                     \hline
Amazon & 47.16\% & 41.30\% & 57.74\% \\
Enron & 39.62\% & 29.49\% & 26.04\% \\
%Clean transfer rate       & 92.80\%    & 93.60\%    & 92.60\% \\
\bottomrule   
\end{tabular}}
\label{tab:insights3}
\vspace{-3mm}
\end{table}

\subsection{Defense Overview}
\begin{figure}[t] 
    \centering
    \includegraphics[trim={8.1cm 5cm 5cm 8cm},clip, scale=0.54]{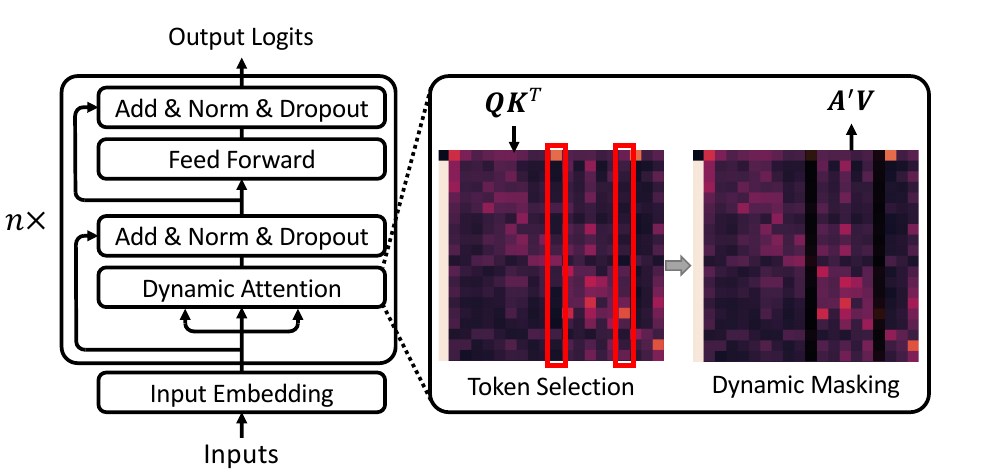}
    \caption{The overview of dynamic attention.}
    \label{fig:overview}
    \vspace{-5mm}
\end{figure}

Based on our discoveries in Sec. \ref{sec:design_intuition}, we propose the dynamic attention mechanism to 1) mitigate the effect of mistakenly assigning higher attention values to irrelevant tokens that lead to the model's misclassification; 2) stir up the instability of adversarial examples by model dynamization. 
Therefore, in our defense method, we mask or weaken the attention of tokens with higher attention values. 
This process prevents the model from paying too much attention to the attentive tokens. 
Even if the attention is completely masked, the residual structure can still preserve information from the previous state.
For these selected tokens, masking can weaken their influence on other tokens, whether erroneous or correct.
Unlike dropout, which randomly drops neurons on intermediate representation by setting their value to 0, dynamic attention follows the intuition that incorrect tokens are assigned with high attention values to reduce the influence of several tokens by decreasing their attention values.
To add dynamization to this process, we change the number of masked tokens in each layer and each time they run.
Furthermore, the dynamic attention mechanism can work simultaneously with the dropout or any other modules to further improve the model's robustness. 
The overview of dynamic attention is shown in Fig. \ref{fig:overview}.

%\jiang{ablatiob study}

\subsection{Dynamic Attention}
\paragraphbe{Attention Rectification}
Since the transformer architecture consists of an encoder and a decoder, we use the encoder part to illustrate the dynamic attention mechanism. 
Here, the encoder first maps an input sequence of tokens $(x_1, x_2, \cdots, x_n)$ to continuous token representations $\textbf{z}=(z_1, z_2, \cdots, z_n)$. 
Then, the representations $\textbf{z}$ are projected into the query matrix $\textit{Q}$, the key matrix $\textit{K}$, and the value matrix $\textit{V}$.
In a common transformer-based model, the query, key, and value matrices are projected multiple times and perform dot-product attention in parallel, called multi-head attention. 
For our dynamic attention, we first calculate the attention matrix of each head
$$A_t=\text{softmax}\left(\frac{Q_tK_t^T}{\sqrt{d}}\right)$$
%$$A=\sum_t{\text{softmax}\left(\frac{Q_tK_t^T}{\sqrt{d}}\right)}$$
where $Q_t$ and $K_t$ are the query matrix and key matrix of the $t^{\text{th}}$ head, and $d$ is the dimension of queries and keys.
Then, we sum up the attention map of all heads to get the global attention $A=\sum{A_t}$. 
Each value in the global attention $A$, denoted as $A[i,j]$, represents the total attention of a query token $i$ on a key token $j$. 
Next, we calculate the total attention value of each key token $A_s$ (i.e., sum across query tokens), which reflects the attention that token receives and is defined as
$A_s=\sum_i A[i,j]$.

\SLJ{We first organize the average attention values of all key tokens, arranging them in descending order. The process of token selection differs based on the type of task:
for text classification tasks, we select the top $m$ tokens, as altered words are commonly assigned with higher attention values. For text generation tasks, we keep the top $m_a$ tokens unchanged, and few subsequent tokens are selected, specifically from top $m_a$ to top $m_b$ tokens, as adversarial attacks on generation tasks tend to alter less significant words.
These statements can be justified in experiments in Sec. \ref{sec:nmt} and \ref{sec:sensitivity}. 
%These selected tokens' index form a set denoted as $\mathcal{T}$.
Note that, when selecting tokens, we exclude special tokens, such as {\fontfamily{qcr}\selectfont[CLS]}, {\fontfamily{qcr}\selectfont[SEP]}, {\fontfamily{qcr}\selectfont[MASK]} in BERT and prefix tokens in prefix-tuning and prompt-tuning.
For ease of representation, the indices of these selected tokens form a set, denoted as $\mathcal{T}$. 
For each of these key tokens in $\mathcal{T}$, we adjust the attention value by a reduction factor $\beta$ to obtain a rectified attention map $A'$, defined as }
\begin{align*}
A_t'[i,j]=\left\{\begin{matrix}
A_t[i,j] & j\notin\mathcal{T}\\ 
\beta \cdot A_t[i,j] &  j\in\mathcal{T}
\end{matrix}\right..
\end{align*}

%$$\mathcal{T}=\arg\max_m\left(A_s\right)$$

The $\beta$ can be set to $[0,1]$ where $\beta=0$ indicates the attention value is completely masked, and $\beta=1$ indicates the attention value is not modified.
Finally, we multiply $A_t'$ with the value matrix $V_t$ to obtain an improved output for each head $t$. 
These outputs are concatenated and projected to form the hidden representations of tokens $H=\underset{t}{\text{Concat}}(A_t'\cdot V_t)\cdot W$ and then passed into the next module.
%\begin{align*}
%H&=\underset{t}{\text{Concat}}(A_t'\cdot V_t)\cdot W.
%\end{align*}
%\SLJ{For different layers, the value $m$ in a text classification model or $m_a$ and $m_b$ in a text generation model can be different.}

\begin{figure*}[t] 
    \centering
    \includegraphics[scale=0.34, trim={50.9cm 0 0 0},clip]{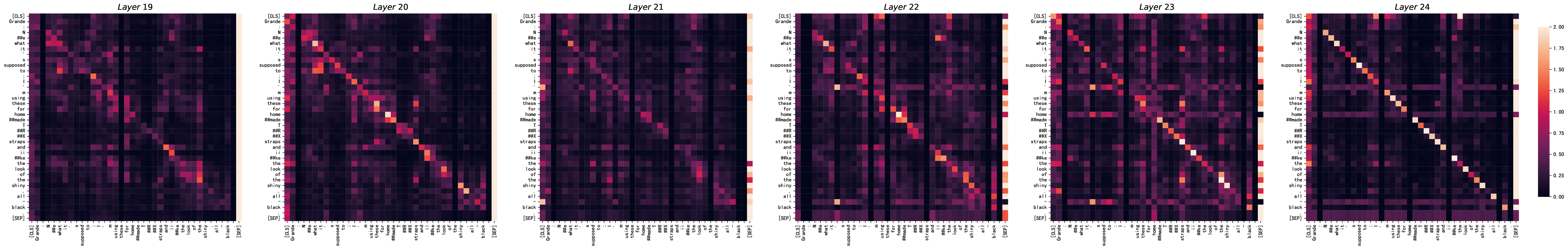}
    \caption{\SLJ{The dynamic attention map of the adversarial example {\em No. 40}.}}
    \label{fig:examples_dynamic}
    \vspace{-7mm}
\end{figure*}

\paragraphbe{Dynamic Modeling}
\SLJ{During attention rectification, we obtain a token index set $\mathcal{T}$ for token weakening or masking. 
However, if the index set $\mathcal{T}$ remains static each time they are run, the model remains vulnerable to attacks, albeit requiring marginally more queries, as adversarial attacks will exploit tokens with lower attention values. 
Therefore, we introduce dynamic modeling into attention rectification, converting the value $m$ into a random variable in text classification tasks, and denoted as $m_i$ for $i^{\text{th}}$ layer.
For each input text, $m_i$ is a random integer drawn from a predefined range proportional to the text's length, given that texts of varied lengths exhibit different numbers of perturbation words.
%As texts of different lengths have different numbers of perturbation words, we base the range on the input text length. 
%Therefore, we determine the range to be a proportion of input text length.
For instance, setting $m_i$'s range between 10\% and 20\% of text length implies that the value of $m_i$ is an integer selected from the interval $[\lfloor 0.1\times n \rfloor, \lfloor 0.2\times n \rfloor]$, with $n$ representing the word or token count in the text, that is,
$m_i\sim\text{discrete\_uniform}(\lfloor 0.1\times n \rfloor, \lfloor 0.2\times n \rfloor).$
Thus, the value of $m$ will change in each run and in each layer.}

\SLJ{In text generation tasks, for each input text, the value of $m_a$ is fixed, while $m_b$ is the random variable that changes across layers and runs. 
We present a sensitivity analysis on $m$ (as well as $m_a$ and $m_b$) and $\beta$ in Sec. \ref{sec:sensitivity} and Appendix \ref{appendix:sensitivity}. }
%According to the experiment of classification tasks studied in Sec. \ref{sec:design_intuition}, where attacks usually modify the task-related words, we commonly make changes to the top few tokens. 
%In addition, shorter texts should use a relatively small range of $m$ intervals, such as the Twitter dataset in Sec. \ref{sec:other_dataset}, where the range of $m$ can be chosen from 1 to 5. 
%For relatively longer texts, a wider range of intervals can be selected. For example, in the Amazon dataset in Sec. \ref{sec:case1}, \ref{sec:case2} and \ref{sec:case3}, we usually choose $m$ in the range of 3 to 8, which is more appropriate.
%For text generation tasks like translation and summarization, we keep the top few tokens unchanged and modify the attention of the latter few tokens. 
%Adversarial attacks toward text generation tasks tend to change the similarity of the generated text to the reference text. 
%Then, the attacks will modify the words that are not critical to understanding the text yet may mislead the model.
%Therefore, the attention of tokens with high attention values (e.g., tokens that are critical to the understanding of the text) is kept unchanged, and the attention of tokens with lower attention (e.g., tokens that are dispensable for model understanding) is weakened slightly.

Finally, dynamic attention can work with other dynamic modeling methods like dropout or any other robustness enhancement techniques like adversarial training and information bottleneck. 
In our experiments, we fuse the dropout, information bottleneck and adversarial training with dynamic attention to further mitigate the impact of adversarial examples.

\subsection{\SLJ{Toy Example}}
\SLJ{To better illustrate the process of dynamic attention, we consider a fine-tuned BERT model and an adversarial text with index {\em No. 40} from the Amazon dataset: ``\textit{[Grande]: [Ne] what it's supposed to; I'm using these for homemade TRX straps and [iike] the look of the shiny, all-black.}'' as an example. 
The corresponding original text is ``\textit{[Great]: [Does] what it's supposed to; I'm using these for homemade TRX straps and [love] the look of the shiny, all-black.}'' 
Words in brackets indicate modifications.}

\SLJ{Given the provided adversarial example with 36 tokens, the range for $m_i$ is determined as $[\lfloor 0.1\times 36 \rfloor, \lfloor 0.2\times 36 \rfloor]=[3, 7]$, i.e., $m_i\in \{3,4,5,6,7\}$. 
As we move forward in the model, these tokens are mapped into token embeddings, which are fed into the first layer.
For the first layer, $m_1$ is randomly selected from the range, and in our case, it is chosen as 3.
Then, the three tokens with the highest attention values will be masked; that is, their attention values are set to 0.
%The process is iterative, with the embeddings being sequentially passed to the succeeding layers. 
%For each layer, the value of $m_i$ is independently and randomly chosen from our established range.
This process is repeated for subsequent layers, where each $m_i$ is independently and randomly chosen from the range.
}

When we reach {\em Layer 22}, as illustrated in Fig. \ref{fig:examples_dynamic}, the model opts for seven tokens, signifying $m_{22}=7$, which are `Grande', `:', `'', `these', `straps', `-', and `.'. 
In this layer, the token `Grande' stands out due to its heightened attention value. 
However, the dynamic attention mechanism steps in, masking this word alongside others, thereby diluting its semantic influence in the overall computation.
The narrative continues in {\em Layer 24}, where six tokens are highlighted, corresponding to \(m_{24}=6\), which are `:', `'', `home', `ii', `-', and `.'.
In this layer, the token `ii' contains significant attention, yet again, dynamic attention ensures it is masked. 
Consequently, adversarial words are masked in Layers 22 and 24, safeguarding the output representation from potential misclassifications. 
Finally, the dynamic attention model can classify the adversarial examples correctly.

%-------------------------------------------------------------------------------
\section{Experimental Settings}
%-------------------------------------------------------------------------------

\subsection{Datasets}

Our method is evaluated on two popular tasks: classification and text generation. 
For the classification task, we use {\em Amazon} (sentiment analysis), consisting of 3.6m product reviews marked as positive or negative \cite{mcauley2013hidden}; {\em Twitter} (toxic comment detection), consisting of 77k tweets annotated as toxic or non-toxic \cite{wijesiriwardene2020alone}; and {\em Enron} (spam detection), consisting of 28k emails marked as spam or non-spam \cite{metsis2006spam}. Additionally, we explore the impact of dataset shift on our method's performance by evaluating it on the {\em Yelp} dataset \cite{zhang2015character}, which contains 560k business reviews marked as positive or negative.
For the text generation task, we evaluate our method on the {\em TED Talk} dataset \cite{cettolo-etal-2012-wit3}, which contains the original Ted talks and their translations in more than 109+ languages, and the {\em Gigaword} dataset \cite{graff2003english}, which contains headlines generated from a corpus of approximately 4 million articles.

\subsection{Baselines}\label{baseline}
We evaluate the effectiveness of dynamic attention by comparing it to four baselines: \mct{i} {\em No defense (Original )}, which denotes the native language model without dynamic modules; 
% \mct{ii} {\em FreeLB}, which uses adversarial training during fine-tuning to enhance model robustness; 
\mct{ii} {\em Dropout} \cite{wang2018defensive}, which uses dropout at test time to achieve strong defense effects;
\mct{iii} {\em Empirical Adversarial Training (AT)}, where we use A2T \cite{yoo2021towards} to enhance model's robustness by training against adversarial examples;
\mct{iv} {\em Information Bottleneck (IB)} \cite{zhang2022improving}, which is an information-bottleneck-based approach to improve the adversarial robustness of a language model.

\subsection{Attack Method}
Since word-level attacks are more challenging for both humans and DNNs to detect~\cite{du2021combating, wang2020defense, wang2022better}, we mainly focus on three attacks containing word-level attack, including \mct{i} {\em TextBugger}, which contains both word-level and character-level perturbations by altering letters or replacing words \cite{li2019textbugger}; \mct{ii} {\em TextFooler}, which identifies and replaces important words with synonyms \cite{jin2020bert}; \mct{iii} {\em PWWS}, which replaces words with synonyms based on counter-fitted word vectors \cite{ren2019generating}. 
%For details about the three attacks, please refer to Appendix \ref{appendix:3}.

\subsection{Metrics}
Following previous works \cite{zhang2022improving, liu2022flooding},  we employ three metrics to evaluate the effectiveness of our method. 

\paragraphbe{Attack Success Rate ({\em ASR})} ASR measures the proportion of misclassified adversarial examples, with a higher ASR indicating a stronger attack or weaker defense.

\paragraphbe{Attack Query ({\em Query})} Query measures the average model queries required to generate adversarial texts, where fewer queries indicate higher vulnerability.

\paragraphbe{Bilingual Evaluation Understudy ({\em BLEU})} BLEU measures the correspondence between the reference text and the generated text, with a score ranging from 0 to 1, where a higher score indicates a better defense.

Note that, {\em ASR} and {\em Query} are used in the classification task, while {\em BLEU} is used in the text generation task.

\subsection{Implementation Details}\label{models}
\paragraphbe{Models} We consider three types of transformer architectures: encoder-only, decoder-only, and encoder-decoder. 
For the classification task, we primarily use the cased BERT model \cite{devlin2019}, a popular encoder-only model. The chosen BERT model is the large version, with 24 layers, 1024 hidden units, 16 attention heads, and 336 million parameters. 
The BERT model is tuned in three ways (fine-tuning, prefix-tuning, and prompt-tuning), as introduced in Sec. \ref{sec:lm}. 
We also include GPT-2 as a representative of the decoder-only model for text classification.
Following \cite{radford2018improving}, we add a classification layer on top of the last output representation, which was previously intended for the next token prediction, and fine-tune it for classification.
%The output representation of the last token by GPT-2, previously intended for the next token prediction, is used to perform text classification. 
%It is pre-trained on lower-cased English text. 
For the text generation task, we evaluate the T5-based model \cite{2020t5}, a widely used encoder-decoder model for text generation tasks. 
We set $\beta=0$ and the range of $m$ to 10\% to 20\% of the text length for fine-tuned and prefix-tuned BERT models and $\beta=0.4$ and the range of $m$ to 20\% to 40\% of the text length for prompt-tuned BERT models.
We set $\beta=0.6$, $m_a=0.1$, and the range of $m_b$ to 30\% to 50\% of the text length for the T5 model.
The sensitivity analysis of these hyperparameters is provided in Sec. \ref{sec:sensitivity} and in Appendix \ref{appendix:sensitivity}.

\paragraphbe{Attacks} For the classification task, we consider an attacker who generates adversarial examples that can result in high-confidence misclassifications by the target model. \SLJ{To this end, we set the condition of stopping the attacks as the prediction confidence of the target label greater than 0.6, considering both the diversity of collected adversarial examples and the evaluation using high-confidence adversarial examples \cite{pang2018towards}.} For the text generation task, we modify the goal function of the evaluated attacks as minimizing the BLEU score but keeping the word transformation methods unchanged. \SLJ{For each experiment, 1000 clean text samples are used for evaluation.}

\paragraphbe{Defenses} 
For the dropout model, we set a 10\% dropout rate which is consistent with the training phase. 
\SLJ{In experiments, we enable all dynamic modules in the BERT model and only enable them in the T5 model's encoder according to the results in Appendix \ref{appendix:encoder-decoder}.}
For the AT model, we tune the model using clean samples and adversarial examples generated by A2T \cite{yoo2021towards}.
For the IB model, we use the default setting that contains FreeLB adversarial training to improve the model's robustness.

\begin{table*}[t]
\centering
\caption{The performance of the original model, the dynamic attention model, the dropout model, and the fusion model on three tuning methods and GPT model, as well as IB and AT, under three attacks and three threat models.}
\scalebox{0.87}{
\SLJ{\begin{tabular}{ccccccccccccccccc}
\toprule
\multirow{2}{*}{Tuning}        & \multirow{2}{*}{Model type} & \multirow{2}{*}{$ACC$} & \multicolumn{4}{c}{TextBugger}        &&  \multicolumn{4}{c}{TextFooler}        && \multicolumn{4}{c}{PWWS}              \\ \cline{4-7} \cline{9-12} \cline{14-17}
&                             &                      & $ASR_Q$  & {\em Query}  & $ASR_D$   & $ASR_S$   && $ASR_Q$  & {\em Query}  & $ASR_D$   & $ASR_S$   && $ASR_Q$  & {\em Query}  & $ASR_D$   & $ASR_S$   \\
\hline
\multirow{9}{*}{\begin{tabular}[c]{@{}c@{}}Fine-\\ tuning\end{tabular}}   & original model         & 93.00\% & 38.06\% & 234.44 & 100.00\% & 100.00\% &  & 47.53\% & 379.42 & 100.00\% & 100.00\% &  & 61.29\% & 695.5  & 100.00\% & 100.00\% \\
& dynamic attention      & 93.07\% & 41.85\% & 320.66 & 34.87\%  & 38.42\%  &  & 52.90\% & 650.65 & 24.80\%  & 33.48\%  &  & 55.29\% & 751.88 & 28.52\%  & 32.98\%  \\
& dropout                & 93.20\% & 35.84\% & 398.11 & 26.83\%  & 56.04\%  &  & 45.18\% & 744.54 & 26.30\%  & 46.56\%  &  & 45.06\% & 820.29 & 39.90\%  & 59.44\%  \\
& fusion                 & 92.27\% & 43.91\% & 312.66 & 15.35\%  & 32.20\%  &  & 50.87\% & 656.44 & 12.88\%  & 31.67\%  &  & 50.54\% & 753.89 & 22.75\%  & 33.33\%  \\
\cline{2-17}
& AT                     & 94.60\% & 45.88\% & 181.17 & 100.00\% & 100.00\% &  & 53.70\% & 333.12 & 100.00\% & 100.00\% &  & 60.04\% & 670.08 & 100.00\% & 100.00\% \\
& dynamic attention + AT & 94.53\% & 41.47\% & 314.49 & 44.50\%  & 50.08\%  &  & 55.06\% & 670.92 & 37.55\%  & 45.93\%  &  & 51.81\% & 785.89 & 40.88\%  & 45.77\%  \\
\cline{2-17}
 & IB                     & 95.07\% & 39.16\% & 323.03 & 28.10\% & 42.94\% &  & 49.68\% & 693.89 & 68.82\% & 33.48\% &  & 53.89\% & 773.63 & 82.42\% & 39.65\% \\
 & dynamic attention + IB & 94.00\% & 38.89\% & 324.76 & 26.37\% & 38.98\% &  & 48.31\% & 708.99 & 27.19\% & 29.41\% &  & 45.96\% & 798.21 & 36.11\% & 31.93\% \\
 & fusion +IB             & 93.00\% & 42.21\% & 310.23 & 29.23\% & 30.51\% &  & 52.48\% & 639.44 & 19.75\% & 28.96\% &  & 48.72\% & 765.69 & 25.76\% & 28.07\% \\
\hline
\multirow{6}{*}{\begin{tabular}[c]{@{}c@{}}Prefix-\\ tuning\end{tabular}} & original model         & 94.80\% & 66.67\% & 178.47 & 100.00\% & 100.00\% &  & 78.48\% & 270.12 & 100.00\% & 100.00\% &  & 88.19\% & 632.18 & 100.00\% & 100.00\% \\
& dynamic attention      & 93.40\% & 48.83\% & 297.15 & 40.17\%  & 43.67\%  &  & 67.31\% & 575.52 & 40.26\%  & 38.71\%  &  & 63.95\% & 746.26 & 41.28\%  & 43.06\%  \\
& dropout                & 94.60\% & 47.57\% & 365.62 & 27.33\%  & 52.54\%  &  & 53.18\% & 683.66 & 23.69\%  & 56.19\%  &  & 55.20\% & 800.07 & 37.68\%  & 60.05\%  \\
& fusion                 & 93.13\% & 50.54\% & 289.99 & 16.38\%  & 39.24\%  &  & 59.91\% & 590.16 & 17.99\%  & 40.05\%  &  & 58.79\% & 737.54 & 18.45\%  & 39.95\%  \\
\cline{2-17}
& AT                     & 92.80\% & 63.15\% & 154.19 & 100.00\% & 100.00\% &  & 75.43\% & 263.64 & 100.00\% & 100.00\% &  & 82.54\% & 613.92 & 100.00\% & 100.00\% \\
& dynamic attention + AT & 93.47\% & 46.79\% & 285.05 & 49.47\%  & 45.28\%  &  & 63.54\% & 567.18 & 47.20\%  & 43.33\%  &  & 56.56\% & 776.07 & 52.60\%  & 48.91\%  \\
\hline
\multirow{6}{*}{\begin{tabular}[c]{@{}c@{}}Prompt-\\ tuning\end{tabular}} & original model    & 90.60\% & 63.13\% & 196.89 & 100.00\% & 100.00\% &  & 77.70\% & 295.5  & 100.00\% & 100.00\% &  & 84.55\% & 641.49 & 100.00\% & 100.00\% \\
& dynamic attention & 90.20\% & 50.66\% & 251.23 & 62.88\%  & 50.00\%  &  & 65.63\% & 530.99 & 51.35\%  & 43.18\%  &  & 64.22\% & 724.69 & 61.25\%  & 43.34\%  \\
& dropout           & 91.60\% & 64.54\% & 227.38 & 15.70\%  & 50.70\%  &  & 75.22\% & 434.58 & 15.16\%  & 47.73\%  &  & 71.24\% & 693.28 & 19.57\%  & 52.22\%  \\
& fusion            & 87.60\% & 67.87\% & 218.05 & 18.33\%  & 43.36\%  &  & 75.87\% & 391.35 & 12.84\%  & 37.78\%  &  & 71.88\% & 636.13 & 16.72\%  & 39.69\% \\
\cline{2-17}
& AT                     & 89.60\% & 77.01\% & 131.89 & 100.00\% & 100.00\% &  & 90.40\% & 223.69 & 100.00\% & 100.00\% &  & 92.41\% & 590.29 & 100.00\% & 100.00\% \\
& dynamic attention + AT & 91.27\% & 67.40\% & 202.55 & 57.79\%  & 66.38\%  &  & 81.15\% & 420.48 & 57.47\%  & 67.00\%  &  & 75.49\% & 681.18 & 67.05\%  & 67.08\% \\
\hline
\multirow{4}{*}{GPT}                                                     & original model              & 93.60\%              & 60.47\% & 149.15 & 100.00\% & 100.00\% && 75.85\% & 247.9  & 100.00\% & 100.00\% && 83.12\% & 612.29 & 100.00\% & 100.00\% \\
& dynamic attention           & 93.20\%              & 62.85\% & 245.72 & 41.58\%  & 45.23\%  && 78.74\% & 440.79 & 20.61\%  & 40.85\%  && 68.03\% & 705.19 & 27.63\%  & 50.90\%  \\
& dropout                     & 92.40\%              & 47.73\% & 311.85 & 17.65\%  & 46.29\%  && 54.76\% & 645.69 & 14.49\%  & 42.63\%  && 51.91\% & 771.15 & 23.81\%  & 48.76\%  \\
& fusion                      & 90.90\%              & 47.80\% & 313.19 & 18.43\%  & 40.75\%  && 58.24\% & 603.09 & 12.83\%  & 35.31\%  && 50.88\% & 758.71 & 16.67\%  & 36.50\% 
    \\
\bottomrule
\end{tabular}}} 
\label{table:case12}
\vspace{-5mm}
\end{table*}

\subsection{Threat Model} \label{sec:threat}
Based on the attacker's capability and knowledge, we consider three threat models model providers will encounter.

\paragraphbe{Attacker's Objective} In classification tasks, the attacker's objective is to manipulate inputs so that the target model classifies them incorrectly. In text generation tasks, the attacker aims to generate texts significantly different from the reference texts, measured by the BLEU score.

\paragraphbe{Attacker's Capability} In this work, we investigate three adversarial attack settings with strong attacker capabilities: 
\mct{i} {\em Query Attack}, the attacker iteratively queries the target model and updates the sample texts based on the returned logits, and once the target model misclassifies it, the attack is successful;
\mct{ii} {\em Dynamic Transfer Attack}, where the attacker has no access to the target model and uses a surrogate model with the dynamic attention mechanism (e.g., the identical model's API or open-source dynamic models) to generate adversarial texts and perform transfer attack; 
\mct{iii} {\em Static Transfer Attack}, where the attacker has no access to the target model and uses a static surrogate model (e.g., open-source static models) to generate adversarial texts and perform transfer attack.
%\mct{iv} {\em Adaptive Attack}, where the attacker is aware of the mechanism of the dynamic attention and generate adversarial examples with desired attention weights.

% \SLJ{For the first setting, the attacker can directly access the target model's output and generate adversarial examples by iteratively querying the model. 
% For the second setting, the attacker cannot directly access the online targe model and can only use a surrogate model to generate adversarial examples, and use transfer attacks to attack the target model.}

\paragraphbe{Attacker's Knowledge}
In our experiments, 
we consider that, in the {\em Query Attack}, the attacker has the output logits or confidence of the target model;
in the {\em Dynamic Transfer Attack}, the attacker has the identical dynamic model locally as the target model (or official API) to perform adversarial attacks;
in the {\em Static Transfer Attack}, the attacker has a static model with the same parameters as the target model but without dynamic attention or other dynamic modules enabled.

\section{Evaluation of Effectiveness}\label{sec:effectiveness}
We first use the texts from the Amazon dataset to illustrate the effectiveness of the dynamic attention under the three threat models stated in Sec. \ref{sec:threat}.

\subsection{Query Attack}\label{sec:case1}
\SLJ{Under the first threat model, where the attacker can access the prediction confidence of the target dynamic online model, we report the clean accuracy ($ACC$ column) of the models, as well as the average ASR ($ASR_Q$ column) and the average attack queries ($Query$ column) for the three adversarial attack methods on the three tuning methods and GPT model in Table \ref{table:case12}. }
For each tuning method, we have four models: the original static model, the dynamic attention model, the dropout model, where only dropout is enabled, and the fusion model, where both dynamic attention and dropout are enabled.

Table \ref{table:case12} reveals that the clean accuracy of the four models under three tuning methods is similar, except for a slight drop in the fusion model. 
\SLJ{Interestingly, dropout models occasionally surpass the original ones, likely due to the fact that it remains consistent with the training process.
However, without dynamic attention in the training phase, its clean accuracy and that of the fusion model drop slightly.
Next, we observe that the model with dynamic attention only decreases the ASR in several scenarios, which is due to a smaller robustness space we justify in Sec. \ref{sec:analysis}. 
Nevertheless, it significantly increases the query number for generating adversarial examples. 
Such an increase in attack queries should be more noticeable for the first threat model, which could bolster query-based black-box defenses \cite{li2022blacklight, byun2022effectiveness, qin2021random} that detect highly similar queries in the input space. 
%For the fine-tuned model, the attack queries of TextBugger increased from 234 to 339, and that of TextFooler increased from 379 to 607.
%Such a significant increase can help query-based black-box defenses achieve better performance.
Similar results can be found in prefix-tuned, prompt-tuned, and GPT models.}

\SLJ{Generally, the dropout model performs slightly better than the dynamic attention model, with the former showing a lower $ASR_Q$ and slightly higher attack queries. 
Moreover, we observe that the fusion and dropout models show similar performance, but dropout suffers from the problem of unstable predictions, which is easier to bypass with multiple trials, as we detail in Sec. \ref{sec:stable}.}

\SLJ{We also find that the ASRs drop notably in the prefix-tuned model but marginally in the fine-tuned and prompt-tuned models.
Overall, fine-tuned models demonstrate superior robustness regarding ASR and attack queries. 
Hence, fine-tuning is preferable due to its lower ASR and larger attack queries. 
However, the costliness of fine-tuning makes prefix-tuning more suitable, which offers lower ASR and higher attack queries that approximate fine-tuning's result and also achieve high clean accuracy.}

\subsection{Dynamic Transfer Attack}\label{sec:case2}
\SLJ{Since online model providers often provide their models as a service without disclosing the prediction confidence, attackers are typically required to generate adversarial texts using a local surrogate model to launch transfer attacks on the online model. 
Addressing the second threat model, where attackers only have access to a local dynamic surrogate model's output, we record the average ASR of locally generated adversarial texts on the target dynamic model.
For consistency, the same dynamic model used online in Sec. \ref{sec:case1} is employed locally.}
Specifically, we use the adversarial texts generated in the first threat model to attack the target model again and record the ASR.
The results of transfer ASRs are presented side-by-side with the result of the query attack, shown in the $ASR_D$ column in Table \ref{table:case12}.

The table reveals that adversarial texts are all misclassified by the original model since the original model is static.
Next, we can see that over half of the adversarial texts generated from the dynamic attention model fail to attack the online model. 
For instance, while 52.90\% of the clean texts become adversarial examples via TextFooler (shown in the $ASR_Q$ column), only 24.80\% can re-attack the target model (shown in the $ASR_D$ column).
This discrepancy arises from the dynamic model's varying states during adversarial example generation and online deployment, reducing adversarial text efficacy.
%This is because the model's dynamic state when adversarial texts are generated is inconsistent with the state when adversarial texts are deployed online, making many adversarial texts lose efficacy. 
%This result is consistent with our third intuition, which verifies that the low transferability of adversarial texts can be used to mitigate the effect of adversarial attacks.
Further, the dropout model records a lower ASR than the dynamic attention model in most cases. 
%For example, in fine-tuned models, the ASR of adversarial texts from TextFooler is 32.40\% under the dynamic attention model but only 26.30\% under the dropout model.
This is because dropout greatly obfuscates the gradients used in generating adversarial examples.
Specifically, due to the difference in neurons being dropped, the adversarial texts generated in the previous dropout state may not influence the neurons in the current dropout state and fail to attack successfully during deployment.
%However, Sec. \ref{sec:stable} reveals that the prediction of the dropout model is unstable, Sec. \ref{sec:analysis} show that the dropout model has a smaller robustness space.
Integrating dynamic attention and dropout, the fusion model further reduces the TextFooler's ASR from 26.30\% to 12.88\%. 
Similar observations can be found in other models and attack methods.

In conclusion, dropout introduces a higher degree of randomness to obfuscate the generated examples, leading to a lower ASR. 
With dynamic attention and dropout working together, the fusion model can further decrease the ASR.

\subsection{Static Transfer Attack}\label{sec:case3}
Under the third threat model, where the attacker only has a static model locally, we consider that the attacker obtains the static version of the target dynamic model to generate adversarial texts and tests them on the target model.
The experiment involves applying three attack methods to the static model, using the Amazon dataset.
Then, the generated adversarial texts are fed into dynamic attention models, dropout models, and fusion models. 
Specifically, we use adversarial texts generated from the original model to attack the other three dynamic models.
We report the attack success rate of locally generated adversarial texts in the $ASR_S$ column of Table \ref{table:case12}.
Since we use adversarial texts from the original model to attack the target model, the attack success rate for the original model is 100.0\%.
%Note that the clean accuracy of the original static model 

Table \ref{table:case12} shows that the adversarial examples that are all misclassified in the original model can be partially classified correctly. 
For instance, in fine-tuned models, 33.48\% of adversarial texts generated using TextFooler can bypass the dynamic attention model, which means about 66\% of these adversarial examples are classified correctly. 
Similarly, in the dropout model, these adversarial texts achieve a 46.56\% ASR, which is higher than the dynamic attention model.
When dynamic attention and dropout are combined, ASR is only 31.67\%. 
Thus, the fusion models achieve the lowest ASR. 
It can be seen that the dynamic attention model outperforms the dropout model in most scenarios of the third threat model. 
In the fine-tuned and prefix-tuned models, the fusion model has achieved the best performance. 
%However, in the prompt-tuned models, the dynamic attention model has achieved the best performance.
Nevertheless, we have shown that dropout introduces excessive randomness which will be easily bypassed by multiple trials in Sec. \ref{sec:stable}.
Additionally, to underscore the versatility of dynamic attention, we experiment on adversarial examples under varying confidence levels in Appendix \ref{appendix:varying_confidence}.

\subsection{Comparison with information bottleneck and adversarial training}
In this section, we compare the performance of dynamic attention with information bottleneck (IB) and adversarial training (AT) and show that dynamic attention can work together with them to improve the model's robustness. 
The result is shown in Table \ref{table:case12}. 

For AT, it fails to improve the model's robustness in most cases and even reduces attack queries, making it easier to generate adversarial samples. 
One of the reasons is that the attack algorithm used in adversarial training is inconsistent with the attack algorithm used in actual attacks. 
However, adversarial training cannot traverse all types of attacks, thus greatly limiting its protection capabilities.
Dynamic attention can be integrated with an adversarially trained model to improve its robustness by reducing the ASR and increasing the number of queries.
For information bottleneck, it performs better than the dynamic attention model under the first threat model, yet dynamic attention outperforms IB under the second and third threat models.
Nevertheless, IB can help the model maintain a high prediction accuracy on clean texts. 
Finally, integrating IB with dynamic attention models and fusion models further improves their robustness. 

\begin{table*}[t]
\centering
\caption{The performance of the original model, the dynamic attention model, the dropout model, and the fusion model on Twitter and Enron.}
\SLJ{\scalebox{0.89}{
\begin{tabular}{cccccccccccccccccc}
\toprule
\multirow{2}{*}{Dataset} & \multirow{2}{*}{Model type} & \multirow{2}{*}{$ACC$} & \multicolumn{4}{c}{TextBugger}          &  & \multicolumn{4}{c}{TextFooler}          &  & \multicolumn{4}{c}{PWWS}                \\ \cline{4-7} \cline{9-12} \cline{14-17}
                         &                             &                        & $ASR_Q$ & $Query$ & $ASR_D$  & $ASR_S$  &  & $ASR_Q$ & $Query$ & $ASR_D$  & $ASR_S$  &  & $ASR_Q$ & $Query$ & $ASR_D$  & $ASR_S$  \\ \hline
\multirow{4}{*}{Twitter} & original          & 93.60\% & 32.05\% & 52.19 & 100.00\% & 100.00\% &  & 41.67\% & 115.53 & 100.00\% & 100.00\% &  & 54.91\% & 157.59 & 100.00\% & 100.00\% \\
& dynamic attention & 92.13\% & 33.77\% & 64.87 & 64.96\%  & 74.22\%  &  & 45.32\% & 142.14 & 61.38\%  & 62.74\%  &  & 56.18\% & 158.2  & 61.26\%  & 64.72\%  \\
& dropout           & 93.67\% & 36.32\% & 79.48 & 55.29\%  & 74.67\%  &  & 49.15\% & 156.67 & 48.92\%  & 69.57\%  &  & 53.75\% & 168.21 & 61.22\%  & 78.34\%  \\
& fusion            & 91.73\% & 39.66\% & 70.46 & 51.81\%  & 70.89\%  &  & 46.61\% & 152.16 & 42.88\%  & 62.22\%  &  & 52.75\% & 162.47 & 50.42\%  & 57.59\% \\ \hline
\multirow{4}{*}{Enron}   & original          & 98.27\% & 32.25\% & 963.30 & 100.00\% & 100.00\% &  & 44.02\% & 1706.6 & 100.00\% & 100.00\% &  & 43.61\% & 1683.1 & 100.00\% & 100.00\% \\
& dynamic attention & 96.73\% & 13.25\% & 1443.9 & 19.79\%  & 36.06\%  &  & 15.98\% & 2670.4 & 23.93\%  & 37.79\%  &  & 12.92\% & 2102.8 & 16.67\%  & 32.40\%  \\
& dropout           & 98.33\% & 13.21\% & 1472.0 & 24.22\%  & 45.92\%  &  & 14.23\% & 2746.0 & 23.89\%  & 39.18\%  &  & 11.22\% & 2128.9 & 26.74\%  & 37.68\%  \\
& fusion            & 96.20\% & 15.15\% & 1433.5 & 15.53\%  & 33.54\%  &  & 15.38\% & 2653.1 & 11.26\%  & 28.88\%  &  & 13.12\% & 2080.9 & 10.58\%  & 25.74\%  \\ 
\bottomrule
\end{tabular}}}\label{table:otherdata}
\vspace{-5mm}
\end{table*}

\subsection{Other Classification Datasets}\label{sec:other_dataset}
We have also experimented with our dynamic attention on other security-critical datasets: Twitter and Enron.
We fine-tune the two models and conduct experiments under the three threat models discussed in the previous section.
The results are shown in Table \ref{table:otherdata}.

From Table \ref{table:otherdata}, we first observe that on Twitter, when using dynamic models, the attack queries increase, as shown in the $Query$ column, and the ASRs of the locally generated examples have dropped, as shown in the $ASR_D$ and $ASR_S$ columns. 
However, the ASR, when directly attacking the online model, as shown in the $ASR_Q$ column, does not decrease on Twitter. 
This is because using the dynamic modeling technique will inevitably result in a smaller robust space which we have shown in Sec. \ref{sec:analysis}.
Nevertheless, the ASR of dynamic transfer attacks is halved compared to query attacks. 
Similarly, on the Enron dataset, when directly attacking the online model, all three dynamic models can halve the ASR and significantly increase the attack queries. 
\SLJ{Furthermore, the ASRs of locally generated examples drop to 12.46\% on average under the fusion model. 
% When using the fusion model, the attack success rate is only 10.82\% on average.  
At the same time, the classification accuracy of the clean text is 96.2\%, which is only 2\% lower than the original model.}
Among all three dynamic models, the fusion models perform best in mitigating adversarial attacks despite a clean performance drop.
We also observe that when defending adversarial examples from the local static model, the ASRs under the dynamic attention and dropout models are similar, and the fusion model further reduces the ASR. 

In conclusion, dynamic attention proves to be effective in protecting security-related models against attacks.
Furthermore, the fusion model, which combines dynamic attention and dropout, demonstrates superior performance in defending against adversarial attacks. 

\begin{table}[]
\centering
\caption{The stableness of dynamic attention models, dropout models and fusion models in multiple settings.}
\SLJ{\scalebox{1.}{\begin{tabular}{ccccc}
\toprule
Dataset	& Model             & $\sigma_{ADV}$    & $\sigma_{CLEAN}$  & $ASR_M$  \\
\hline
\multirow{3}{*}{\begin{tabular}[c]{@{}c@{}}Amazon\\ (Fine-tuning)\end{tabular}}     & dynamic attention & 0.1040 & 0.0273 & 47.51\% \\
& dropout           & 0.3742 & 0.0292 & 93.21\% \\
& fusion            & 0.1708 & 0.0604 & 55.66\% \\
\hline
\multirow{3}{*}{\begin{tabular}[c]{@{}c@{}}Amazon\\ (Prefix-tuning)\end{tabular}} & dynamic attention & 0.0731 & 0.0242 & 55.38\% \\
& dropout           & 0.2343 & 0.0363 & 98.92\% \\
& fusion            & 0.1730 & 0.0626 & 75.27\% \\
\hline
\multirow{3}{*}{\begin{tabular}[c]{@{}c@{}}Amazon\\ (Prompt-tuning)\end{tabular}}	& dynamic attention & 0.0239 & 0.0158 & 56.38\% \\
& dropout           & 0.1868 & 0.0947 & 79.79\% \\
& fusion            & 0.1830 & 0.1169 & 81.91\% \\ \hline
\multirow{3}{*}{Twitter}        & dynamic attention & 0.1086 & 0.0260 & 81.54\% \\
& dropout           & 0.2657 & 0.0222 & 98.97\% \\
& fusion            & 0.1945 & 0.0463 & 83.59\% \\ \hline
\multirow{3}{*}{Enron}          & dynamic attention & 0.1309 & 0.0250 & 59.45\% \\
& dropout           & 0.2306 & 0.0196 & 68.20\% \\
& fusion            & 0.1751 & 0.0386 & 59.91\% \\
\bottomrule       
\end{tabular}}}\label{tab:stableness}
\vspace{-5mm}
\end{table}

\subsection{Stableness Evaluation}\label{sec:stable}
In this part, we study the stableness of the dynamic attention, dropout, and fusion models. 
Since we introduce dynamic modeling into dynamic attention, an input's output confidence may change each time they run. 
Consequently, while a dynamic modeling method may exhibit a low ASR, it can introduce excessive randomness so that some inputs can be classified correctly this time yet will be misclassified next time.
For the method that introduces too much randomness, as long as the attacker uploads the adversarial example a few more times, there is a high probability that the target model will misclassify it.
To evaluate the stableness of each model, we measure the average standard deviation of the outputs' confidence through multiple trials.
Specifically, for each dynamic model and dataset, we repetitively query the model 100 times for a given input text and calculate the standard deviation of the resulting confidence values. 
Subsequently, we compute the average standard deviation $\sigma$ for adversarial texts generated using TextFooler (denoted as $\sigma_{ADV}$) and clean texts (denoted as $\sigma_{CLEAN}$).
These metrics serve as indicators of the model's stability, with lower standard deviations indicating greater stability.
The results are presented in Table \ref{tab:stableness}.
Additionally, we compute the static transfer ASR of locally generated adversarial examples with multiple uploads shown in the $ASR_M$ column in Table \ref{tab:stableness}.
%Specifically, for each adversarial example, we upload 20 times, and once one is misclassified, we consider this adversarial example attack successful.
Specifically, we conduct ten trials for each adversarial example, and consider the adversarial attack successful if it is misclassified in at least one trial.
\SLJ{Note that the $\sigma$ and $ASR_M$ for the original static model are 0 and 100\%, respectively. }

From Table \ref{tab:stableness}, we can first observe that the average standard deviation of clean samples is exceptionally low.
This finding suggests that the clean samples exhibit stability, as their output confidence remains relatively unchanged across different dynamic statuses. 
Consequently, clean texts will barely be misclassified by the dynamic models. 
The results of adversarial examples demonstrate that dynamic attention models achieve the lowest standard deviation, implying their superior stability compared to dropout and fusion models. 
%Therefore, the dynamic attention models are more stable than the dropout and fusion models.
We conjecture that the dropout module, which randomly drops out 10\% of neurons, results in the loss of critical information necessary for the model understanding. 
%We speculate that the dropout module, which randomly drops out 10\% of neurons, will inevitably preserve adversarial information in several dynamic statuses. 
%Therefore, the dropout models can classify one adversarial example correctly at this time but have a higher probability of misclassifying it next time. 
Therefore, we speculate that dropout models may inadvertently retain adversarial information in certain dynamic statuses, leading to correct classification of an adversarial example at one trial but a higher probability of misclassification in subsequent trial. 
The results of the attack success rate under ten trials, shown in the $ASR_M$ column, also confirmed this point.
The attack success rate increases from 46.56\% (from Table \ref{table:case12} $ASR_S$ under TextFooler) to 93.21\% with multiple uploads in the fine-tuned Amazon model with dropout enabled.
Whereas, in the dynamic attention model, the ASR increases from 33.48\% to only 47.51\%, which is much lower than the ASR of the dropout model.
We can find similar observations in the prefix-tuned and prompt-tuned models and model trained via the Twitter and Enron datasets.
%, where the $ASR_M$ of the dropout model are both 98\%.
%However, the dynamic attention models have a lower standard deviation, which implies that the dynamic attention model's output label is more consistent than the dropout model. 
%For the models trained using Twitter and Enron, we can also observe that the dynamic attention model predicts more stable output than the dropout and fusion models.
Furthermore, the fusion model exhibits improved stability compared to the dropout model. 
This could be attributed to the masking of highly attentive tokens, which distribute textual information and reduces the adversarial information retained during the dropout process.
%In conclusion, the dynamic attention model has a more stable prediction than the other two dynamic models. 
To conclude, the dynamic attention model offers more consistent predictions than the other two dynamic models, making it more practical for stable deployment.
However, dropout introduces excessive randomness, rendering it impractical to defend against adversarial attacks.

In summary, considering the fine-tuned and prefix-tuned BERT model and the GPT model, it is recommended to use the fusion model, which performs best in all three threat models and achieves lower variation compared to the dropout model. 
Considering the prompt-tuned model, the fusion model is recommended for defending against transfer attacks, and the dynamic attention model is recommended for defending against query attacks.

\begin{table}[t]
\centering
\caption{The BLEU score of clean texts and adversarial texts generated from local static T5 model using TED Talk and Gigawords.}
\scalebox{1}{
\SLJ{\begin{tabular}{ccccc}
\toprule
Task                                                                             & Model type   & Clean       & TextBugger & TextFooler \\ \hline
\multirow{4}{*}{English to French} & original model    & 1.0000 & 0.4698     & 0.4807     \\
                               & dynamic attention & 0.8228 & 0.4905     & 0.5194     \\
                               & dropout           & 0.6186 & 0.3977     & 0.3949     \\
                               & fusion model      & 0.6022 & 0.3601     & 0.3983     \\ \hline
\multirow{4}{*}{English to German} & original model    & 1.0000 & 0.3855     & 0.3715     \\
                               & dynamic attention & 0.8823 & 0.4223     & 0.4192     \\
                               & dropout           & 0.6505 & 0.2844     & 0.3426     \\
                               & fusion model      & 0.6469 & 0.2966     & 0.3568     \\ \hline
\multirow{4}{*}{Summarization}    & original model    & 1.0000 & 0.6159     & 0.5344     \\
                               & dynamic attention & 0.8120 & 0.6276     & 0.5765     \\
                               & dropout           & 0.6149 & 0.5008     & 0.4838     \\
                               & fusion model      & 0.5960 & 0.4687     & 0.3861 \\
                       \bottomrule
\end{tabular}}}\label{table:nmt}
\vspace{-5mm}
\end{table}

\begin{table*}[t]
\centering
\caption{The performance of the original model and the three dynamic models on the dataset shift scenario.}
\SLJ{\scalebox{.95}{\begin{tabular}{cccccccccccccccc}
\toprule
\multirow{2}{*}{Model} & \multirow{2}{*}{ACC} & \multicolumn{4}{c}{TextBugger}         &  & \multicolumn{4}{c}{TextFooler}          &  & \multicolumn{4}{c}{PWWS}                \\
\cline{3-6} \cline{8-11} \cline{13-16}
                       &                      & $ASR_Q$  & {\em Query}  & $ASR_D$   & $ASR_S$   && $ASR_Q$  & {\em Query}   & $ASR_D$   & $ASR_S$   && $ASR_Q$  & {\em Query}   & $ASR_D$   & $ASR_S$   \\
                   \hline
original model         & 93.40\%              & 49.25\% & 291.8  & 100.00\% & 100.00\% &  & 60.60\% & 545.24  & 100.00\% & 100.00\% &  & 74.95\% & 1032.4 & 100.00\% & 100.00\% \\
dynamic attention      & 92.93\%              & 43.01\% & 499.59 & 28.60\%  & 36.09\%  &  & 58.71\% & 1014.1 & 23.93\%  & 35.22\%  &  & 61.64\% & 1223.2 & 34.38\%  & 35.24\%  \\
dropout                & 93.80\%              & 41.88\% & 490.98 & 27.22\%  & 50.87\%  &  & 54.43\% & 983.80  & 27.93\%  & 47.71\%  &  & 60.86\% & 1208.6 & 33.85\%  & 59.15\%  \\
fusion                 & 93.20\%              & 46.01\% & 503.49 & 25.72\%  & 33.33\%  &  & 52.29\% & 1055.6 & 14.86\%  & 32.63\%  &  & 57.67\% & 1241.9 & 19.35\%  & 31.52\% 
 \\
\bottomrule
\end{tabular}}}\label{tab:dataset_shift}
\vspace{-3mm}
\end{table*}

\subsection{Neural Machine Translation and Summarization}\label{sec:nmt}
In addition to text classification tasks, we also experiment with text generation tasks, which include neural machine translation and text summarization. 
We utilize a pretrained T5-base model to translate English to French and German using the prompts  `translate English to French:' and `translate English to German:', respectively. 
Meanwhile, we use the same T5 model to summarize articles into headlines using the prompt `summarize:'.
To adapt adversarial attacks for text generation tasks, we modify the objectives of TextBugger and TextFooler, originally designed for classification tasks. 
\SLJ{Specifically, we aim to minimize the BLEU score between the machine-translated text and the reference translation generated from translating or summarizing the clean texts using the static model.}
In addition, adversarial texts are generated from the original static model using texts from the TED Talk and the Gigaword datasets.
Then, we translate the clean and adversarial text using the original static model, and the three dynamic models, respectively, and calculate the BLEU score of it and the reference text.
\SLJ{It is worth noting that, for the dynamic attention model, we only implement dynamic attention in the encoder since enabling dynamic modeling in both encoder and decoder will deteriorate the performance, shown in Appendix \ref{appendix:encoder-decoder}. }
Similarly, for the dropout model, we only enable the dropout in the encoder while disabling the dropout in the decoder.
Since the text generation task relies on the information of all words, adversarial attacks, aiming to minimize the generated text's BLEU score, usually attack non-keywords.
Thus, we let $\mathcal{T}$ contains token indices whose attention value is slightly lower by keeping tokens with the highest attention value unchanged and choose $\beta=0.6$ to weaken their influence.
\SLJ{Specifically, we choose $m_a=0.1$ and set the range of $m_b$ to be $[0.3, 0.5]$.
This choice of hyperparameters is justified in the sensitivity analysis in Appendix \ref{appendix:sensitivity}.
We report the BLEU scores of the generated texts, fed with clean and adversarial texts, for the four models under two translation tasks (in `English to French' and `English to German' rows) and one summarization task (in `Summarization' row) in Table \ref{table:nmt}.}

%The row of the `original model' shows the BLEU score of the generated adversarial examples and reference texts.
The result in Table \ref{table:nmt} demonstrates that the dynamic attention models have improved the translation quality of adversarial texts.
\SLJ{The BLEU score increases from 0.4753 to 0.5050, on average, when utilizing dynamic attention in the English-French translation. 
However, contrary to the results from text classification tasks, the performance of the dropout model has deteriorated to only 0.3834, on average, which is lower than the performance of the original model.
The result of the fusion model shows that the performance does not improve by adding dynamic attention to dropout.}

\SLJ{Similar results can be found in experiments of the English to German translation and the summarization of articles.}
%, the dynamic attention model can slightly improve the summarization task compared with the original model, where the BLEU score increases from 0.6254 to 0.6487, on average. 
%In addition, the dropout model and fusion model can only degrade the performance of the summarization task.
We believe that, unlike the classification task, where the pivot information is concentrated in a few words, the generation task relies on the information of all words in the text, and dropout will cause the hidden representation of the model to lose part of the textual information, thus damaging the effectiveness of text generation.
\SLJ{In addition, our empirical studies found that adversarial examples in text generation tasks are more likely to change the tense of verbs and the adverbs, thereby achieving the purpose of changing the sentence structure, ultimately leading to a significant decrease in BLEU scores.
However, dynamic attention weakens the attention on potentially erroneous tokens instead of completely masking their influence. }
Therefore, dynamic attention can preserve more information than the dropout.

In conclusion, dynamic attention brings more degrees of freedom for adjustment and optimization and can avoid excessive loss of information.
Moreover, dropout is not as good as dynamic attention, which contradicts the results from text classification tasks.
\SLJ{Thus, it is recommended to use the dynamic attention model in text generation tasks.}

\begin{figure}[] 
    \centering
    \includegraphics[scale=0.24]{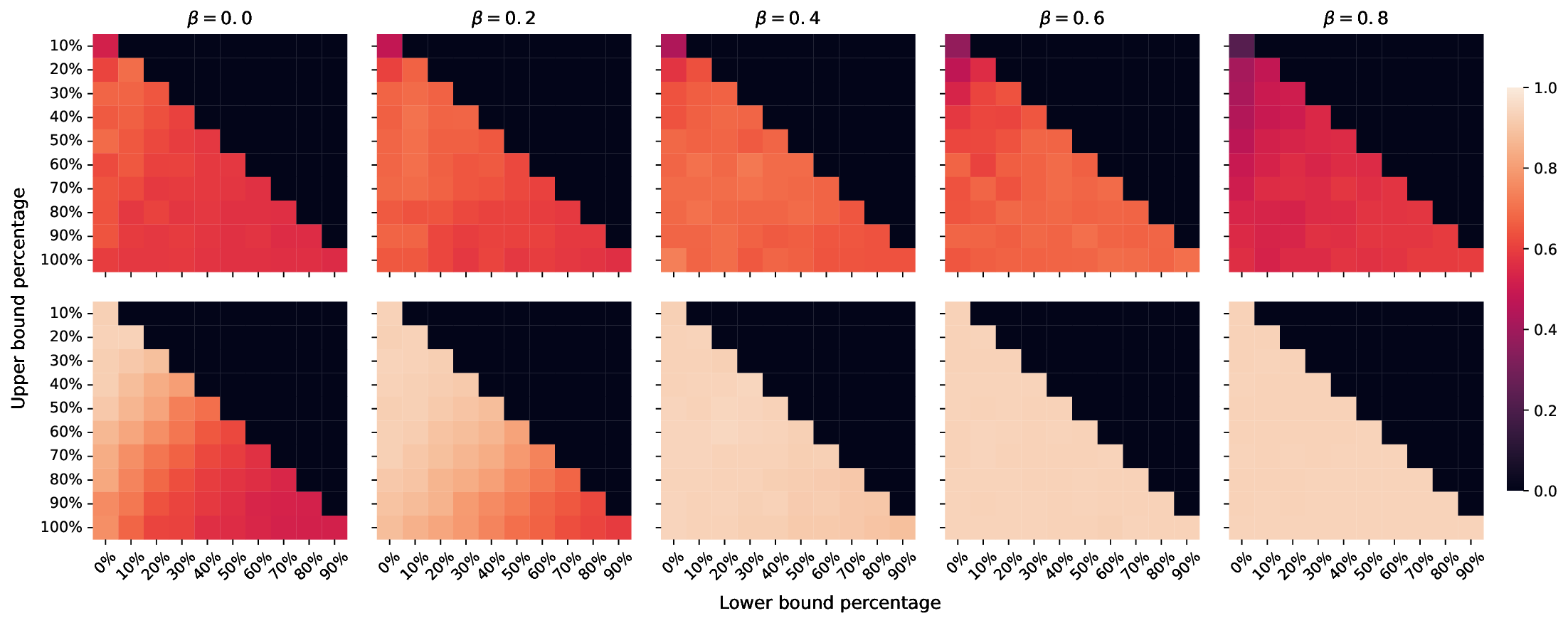}
    \caption{\SLJ{The prediction accuracy of adversarial texts (upper row) and clean texts (bottom row) from the Amazon dataset under the {\em fine-tuned} model with different $m$'s range and $\beta$ value.}}
    \label{fig:amazon_m}
    \vspace{-5mm}
\end{figure}

\begin{figure}[] 
    \centering
    \includegraphics[scale=0.24]{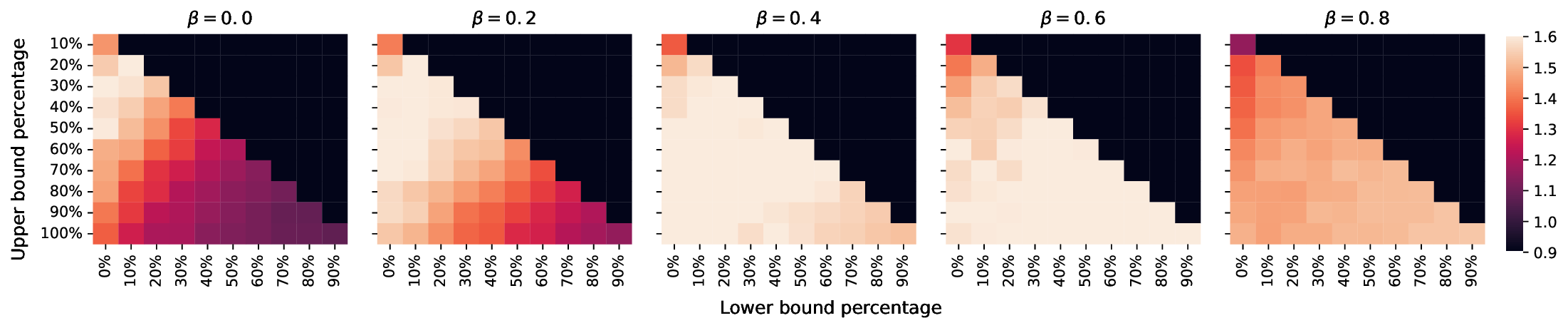}
    \caption{\SLJ{The overall metric $M$ of the dynamic attention model trained with the Amazon dataset under the {\em fine-tuned} model with different $m$'s range and different $\beta$ value.}}
    \label{fig:amazon_b}
    \vspace{-5mm}
\end{figure}

\subsection{Sensitivity Analysis}\label{sec:sensitivity}
In this section, we study how hyperparameter $m$'s range and $\beta$ influence the performance of dynamic attention. 
Firstly, we consider the text classification scenario.
Systematically, the lower bound for the range of $m$ spans from 0\% to 90\% in increments of 10\%, while the upper bound extends from 10\% to 100\%, also in 10\% increments. 
Besides, the value of $\beta$ varies from 0.2 to 0.8 with an increment of 0.2.
Our experiments utilized adversarial texts from the Amazon, generated via a static model, to assess the efficacy of the dynamic attention model across varying $m$'s ranges and $\beta$'s values. 
We provide the prediction accuracy for both adversarial examples ($ACC_a$) and non-adversarial or clean examples ($ACC_o$) under different hyperparameter configurations, as depicted in Fig. \ref{fig:amazon_m}. 
To holistically gauge hyperparameter performance on both adversarial and clean texts, we introduced a composite metric $M=ACC_a+ACC_o$, as shown in Fig. \ref{fig:amazon_b}.
Analyzing the metric $M$, we can observe that a suitable range of $m$ can be determined without setting a smaller upper bound or a larger lower bound. 
For instance, at $\beta=0$, the sweet spot seems to be an upper bound of 20\% and a lower bound of 10\% for peak performance. As $\beta$ escalates, so does the ideal range, implying wider bounds may be more apt. 
We have also included the sensitivity analysis of other models and tasks in Appendix \ref{appendix:sensitivity}.

\section{Evaluation of Versatility}
\subsection{Dataset Shift}
%Previous experiments have shown that dynamic attention is effective in reducing the attack success rate and increasing the attack queries of adversarial attacks and the fusion model which combines dynamic attention and dropout can further improve the defense performance in classification tasks.
Considering that the variation in attentive tokens between the training and evaluation sets can potentially undermine dynamic attention's efficacy, we investigate the effectiveness of dynamic attention in dataset shift scenarios, where the evaluation set differs from the training set. 
Specifically, we use TextBugger, TextFooler, and PWWS to attack the three dynamic models and the static model trained from the Amazon dataset using adversarial texts from the Yelp dataset. 
We perform attacks under the three threat models in Sec. \ref{sec:threat}.
%Then, the generated adversarial examples are used to attack the three dynamic models. 
Then, we report the prediction accuracy of clean texts from the Yelp dataset ($ACC$), the attack success rate ($ASR_Q$), and attack queries ($Query$) when directly attacking an online model, the ASR of adversarial texts generated from the local dynamic model ($ASR_D$) and the ASR of adversarial texts generated from the local static model ($ASR_S$) in Table \ref{tab:dataset_shift}.

We can first observe from the table that the model trained from the Amazon dataset achieves high prediction accuracy on the clean Yelp dataset. 
%Next, the dynamic attention model effectively increases the attack queries of the three attacks and decreases the $ASR_Q$ of Yelp texts generated from the local dynamic model.
Next, the dynamic attention model demonstrates a notable increase in attack queries for all three attacks, leading to a decrease in $ASR_Q$ for the adversarial Yelp texts generated from the local dynamic model.
\SLJ{In the first two threat models, the dynamic attention model and the dropout model have similar performance. }
%However, in these same cases, the fusion model surpasses the dropout model.
\SLJ{In the third threat model, the dynamic attention model outperforms the dropout model, where adversarial Yelp texts are generated from a local static model. }
This indicates that dynamic attention is effective in defending against attacks using shifted datasets. 
\SLJ{Nonetheless, the fusion model achieves the best performance considering both prediction accuracy and ASR under three threat models.} 

%The dropout model slightly outperforms the dynamic attention model in the first two cases.
%Nevertheless, the fusion model outperforms the dropout model in the first two cases. 
%The dynamic attention model outperforms the dropout model under the third threat model, where adversarial Yelp texts are generated from the local static model.
%Thus, dynamic attention is effective in defending against attacks using shifted datasets, yet the fusion model achieves the highest prediction accuracy on adversarial texts.

\begin{figure}[t] 
    \centering
    \includegraphics[scale=0.4]{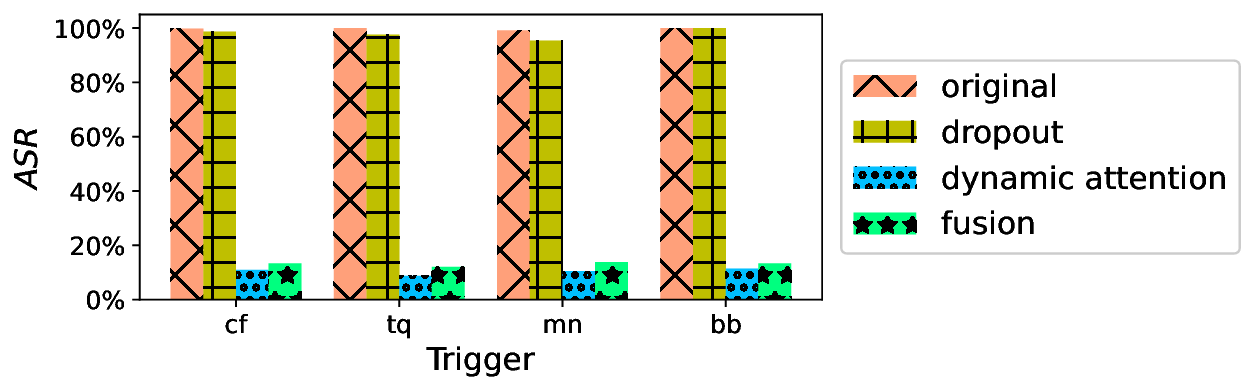}
    \caption{\SLJ{The performance of dynamic models under BadNets attack.}}
    \label{fig:backdoor1}
    \includegraphics[scale=0.48]{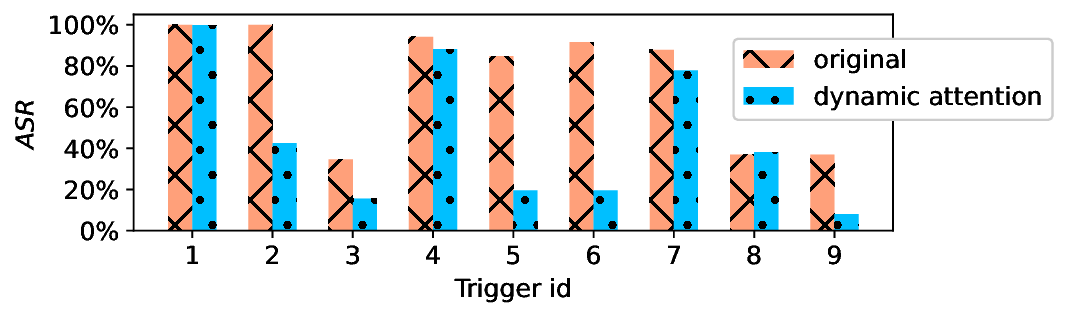}
    \caption{The dynamic model's performance under POR attack.}
    \label{fig:backdoor2}
    \vspace{-5mm}
\end{figure}

\subsection{Backdoor Attack}
In this section, we investigate the effectiveness of dynamic attention in mitigating the impact of backdoor attacks.
\SLJ{We employ two backdoor attacks, BadNets \cite{gu2019badnets} and POR \cite{10.1145/3460120.3485370}.
BadNets poisons the fine-tuning Amazon dataset with specific triggers and target labels using BERT as the target model. 
POR injects nine triggers simultaneously before the model is fine-tuned, and we fine-tuned the backdoor model using the Amazon dataset. 
We evaluate the fine-tuned models and calculate the ASR of these triggers on both the original static model and the dynamic models, as shown in Figs. \ref{fig:backdoor1} and \ref{fig:backdoor2}.}

\SLJ{Fig. \ref{fig:backdoor1} demonstrates that dynamic attention can effectively attenuate the BadNets's backdoor effects, reducing ASR from 100\% to less than 10\%.
In contrast, the dropout model is ineffective in removing the BadNets's effects.
However, Fig. \ref{fig:backdoor2} demonstrates that dynamic attention can reduce the ASRs of several triggers, such as No. 2, 3, 5, 6, and 9. 
We can conclude that the backdoor attacks, which poison the training dataset like BadNets, take effect by allocating high attention to these trigger tokens and thus result in misclassification. 
Therefore, dynamic attention can effectively find these attentive triggers and eliminate their backdoor influence to a large extent.
However, backdoor attacks like POR, which associate triggered texts with target hidden representations, are more elusive and harder to defend.}
%This effectiveness stems from the fact that backdoor triggers manipulate attention by allocating higher attention value to the trigger tokens, thereby influencing the model's output.
%However, high attention tokens, including the backdoor trigger tokens, can be masked with attention rectification, nullifying their impact at the attention layer.
%Thereby, the backdoor effects will be greatly attenuated by dynamic attention.

\begin{table*}[t]
\centering
\caption{The performance of the dynamic attention model under two adaptive attacks with different local model.}  
\SLJ{\scalebox{0.9}{   
\begin{tabular}{cccccccccccccc}
\toprule
&                   & \multicolumn{4}{c}{TextBugger}                        & \multicolumn{4}{c}{TextFooler}                        & \multicolumn{4}{c}{PWWS}                              \\
\cline{3-14}
&                   & $ASR_{SL}$ & $ASR_{ST}$ & $ASR_{DL}$ & $ASR_{DT}$ & $ASR_{SL}$ & $ASR_{ST}$ & $ASR_{DL}$ & $ASR_{DT}$ & $ASR_{SL}$ & $ASR_{ST}$ & $ASR_{DL}$ & $ASR_{DT}$ \\
\hline
\multirow{3}{*}{\begin{tabular}[c]{@{}c@{}}Fine\\ -tuning\end{tabular}}   & dynamic attention & 38.06\% & 37.29\% & 41.85\% & 32.82\% & 47.53\% & 34.24\% & 52.90\% & 22.22\% & 61.29\% & 32.63\% & 55.29\% & 27.21\% \\
& adaptive 1        & 21.29\% & 41.41\% & 21.94\% & 21.24\% & 29.46\% & 37.47\% & 30.11\% & 23.33\% & 31.83\% & 39.41\% & 29.40\% & 32.85\% \\
& adaptive 2        & 4.09\%  & 63.16\% & 7.53\%  & 43.81\% & 6.88\%  & 55.21\% & 9.72\%  & 44.44\% & 14.19\% & 51.01\% & 13.76\% & 34.38\% \\
\hline
\multirow{3}{*}{\begin{tabular}[c]{@{}c@{}}Prefix\\ -tuning\end{tabular}} & dynamic attention & 66.67\% & 47.15\% & 48.83\% & 37.26\% & 78.48\% & 40.32\% & 48.83\% & 38.28\% & 88.19\% & 42.58\% & 63.95\% & 42.62\% \\
& adaptive 1        & 40.51\% & 46.88\% & 22.70\% & 47.80\% & 55.06\% & 43.93\% & 37.23\% & 43.05\% & 63.29\% & 41.44\% & 33.90\% & 47.59\% \\
& adaptive 2        & 19.62\% & 57.35\% & 19.32\% & 42.49\% & 27.22\% & 50.90\% & 28.63\% & 44.78\% & 36.71\% & 52.68\% & 33.69\% & 49.16\% \\
\hline
\multirow{3}{*}{\begin{tabular}[c]{@{}c@{}}Prompt\\ -tuning\end{tabular}} & dynamic attention & 63.13\% & 49.42\% & 50.66\% & 61.57\% & 77.70\% & 42.33\% & 65.63\% & 54.84\% & 84.55\% & 43.52\% & 64.22\% & 61.25\% \\
& adaptive 1        & 39.74\% & 48.89\% & 28.21\% & 30.89\% & 58.94\% & 43.82\% & 42.66\% & 30.65\% & 60.49\% & 45.50\% & 32.27\% & 42.49\% \\
& adaptive 2        & 14.57\% & 66.16\% & 13.53\% & 43.50\% & 20.97\% & 55.09\% & 20.36\% & 39.26\% & 25.39\% & 59.13\% & 22.43\% & 54.76\% \\
\bottomrule   
\end{tabular}}}\label{table:adaptive}
\vspace{-3mm}
\end{table*}

\begin{figure*}[t]
	\centering
    \includegraphics[scale=0.65]{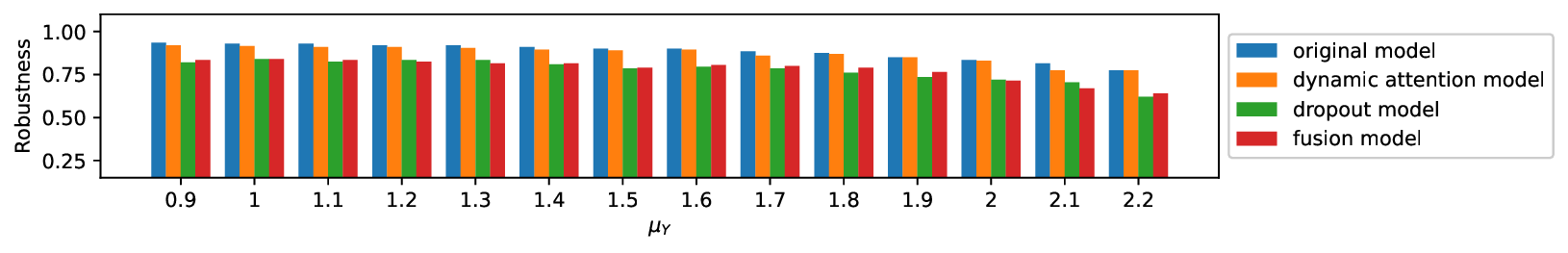}
    \caption{The percentage of robust samples under the attack of 10\% modification rates.}
    \label{fig:robustness_0.1}
    \vspace{-5mm}
\end{figure*}
%-------------------------------------------------------------------------------
\subsection{Possible Adaptive Attack}\label{sec:adaptive}
%-------------------------------------------------------------------------------
\SLJ{We investigate two adaptive attack strategies toward the text classification task, where the attacker uses a local model to generate adversarial texts that satisfy the desired attention weight distribution: (1) assigning high values to relevant tokens (i.e., altering later tokens whose rank index are larger than $m$) and (2) assigning comparable attention values to all tokens.} 
Given that text-based adversarial attacks in the text domain are performed on discrete word space and require the greedy search of feasible words, it is impossible to manipulate the attention weights directly. 
Hence, we constrain the text generated by greedy search to meet the desired attention weights. 
%We find that these adaptive attacks are ineffective in bypassing dynamic attention without significantly lowering the ASRs of generating adversarial texts on local models.

Firstly, assigning high values to relevant tokens of the generated adversarial text helps prevent irrelevant toxic tokens from being removed by dynamic attention.
\SLJ{To achieve this, we enforce a higher overlap between tokens with high attention values in the generated adversarial text ($\mathcal{T}_g$) and tokens with high attention values in the original clean text ($\mathcal{T}_o$), such that: $\frac{|\mathcal{T}_g\cap\mathcal{T}_o|}{|\mathcal{T}_g\cup\mathcal{T}_o|}>0.8$.}
These tokens are selected from the last six layers, with the top five tokens chosen based on their highest attention values in each layer.

Secondly, assigning comparable attention values to all tokens is possible to misguide the token selection process during attention rectification, ensuring that toxic tokens are not removed by dynamic attention.
\SLJ{To achieve this, we enforce the standard deviation of the average attention value $A_s$ of all tokens to be lower than a threshold: $\sigma(A_s)<1.5$.}

We conduct adaptive attacks on local static and dynamic models and perform transfer attacks to the target models. 
%The $ASR_{SL}$ indicates the adaptive ASR under local static model; the $ASR_{ST}$ indicates the transfer ASR using adversarial texts generated from adaptive attack with local static model; the $ASR_{DL}$ indicates the adaptive ASR under local dynamic model; the $ASR_{DT}$ indicates the transfer ASR using adversarial texts generated from adaptive attack with local dynamic model.
We record the ASR on local models using adaptive attacks: $ASR_{SL}$ and $ASR_{DL}$, and the ASR of the transfer attack: $ASR_{ST}$ and $ASR_{DT}$, differentiated by static and dynamic local models respectively. 
The experimental results of models with different tuning methods are shown in Table \ref{table:adaptive}. 
Although the two adaptive attacks yield slightly higher $ASR_{ST}$ on the fine-tuned model, they fail to increase the $ASR_{ST}$ on the prefix-tuned and prompt-tuned models. 
Moreover, to achieve higher transfer ASR, it drastically decreases the local ASR, which means fewer texts can be attacked successfully, even less than the successfully attacked texts without adaptive attack.
\SLJ{In addition, as shown in $ASR_{DL}$ and $ASR_{DT}$, using dynamic local models does not aid the adaptive attacks.}
Thus, dynamic attention is still effective in defending against adaptive attacks.

%-------------------------------------------------------------------------------
\section{Robustness Analysis}\label{sec:analysis}
%-------------------------------------------------------------------------------

% \usepackage{multirow}

%\subsection{Statistical  Analysis}\label{sec:certified}
In this section, we examine the robustness of the dynamic attention model and the other two dynamic models as well as the original model using the Amazon dataset.
Given the inherent difficulty of precisely measuring the certified robustness of a deep Transformer-based model and capturing the robustness space of a dynamic model, we employ a statistical robustness assessment \cite{webbstatistical} and relax the requirement for computing the robustness radius by calculating the percentage of samples that are robust under random noises.

In our approach, we introduce perturbations to test samples by randomly selecting 10\% of the tokens in the token sequence and adding standard Gaussian noise. 
The noise is applied to each dimension of the embeddings of the selected tokens, with a standard deviation $\sigma$.
%Specifically, for each test sample, we randomly select 10\% of the tokens in the token sequence and add standard Gaussian noise with a standard deviation of $\sigma$ to each dimension of the selected tokens' embeddings. 
It's important to note that selecting 10\% of tokens is to maintain consistency with the practical attack \cite{app10103559}.
Additionally, we perform experiments with 20\% and 40\% randomly selected tokens to explore different levels of modification rate.
This process is repeated 500 times, generating 500 noise copies of the test sample.
Then, we examine whether these noise copies of a test sample are all classified correctly by the model.
If there exists one misclassified copy, we consider this test sample not robust to the perturbation strength of $\sigma$.
%In order to correspond to the actual attack scenario, we determine the range of $\sigma$ according to the perturbation size of the actual adversarial attacks.
%We use synonym replacement attacks, such as TextFooler and PWWS, as representatives.
%\SLJ{The distribution of the Euclidean distance between the original words and replaced words under TextFooler and PWWS is shown in Fig. \ref{fig:dist}.
%The figure shows that the distance between the original and replaced words typically falls within the range of 0.9 to 2.2. }
\SLJ{We set the range of $\mu_Y$ from 0.9 to 2.2, which is determined from the actual distance as shown in Appendix \ref{appendix:euclidean}, and choose $\sigma=0.03125\cdot\mu_Y$, as the distance between the noise copies and the original word exhibits a $\chi$-like distribution.}
Thus, we consider 200 texts and test them at 14 perturbation strengths ranging from 0.9 to 2.2. 
The percentage of test samples whose noise copies remain robust to the perturbation is computed, which we define as the statistical robustness of the model. 
It is worth mentioning that a higher percentage value indicates a more robust model.
We report the statistical robustness of the four models with different perturbation strength under a 10\% modification rate in Fig. \ref{fig:robustness_0.1}. 
Additionally, we provide the statistical robustness under 20\% and 40\% modification rates in Figs. \ref{fig:robustness_0.2} and \ref{fig:robustness_0.4} in Appendix \ref{appendix:robustness}, respectively.

\SLJ{From Fig. \ref{fig:robustness_0.1}, we observe that the original model (blue bars) exhibits the highest robustness.
% likely due to the dynamic models's instability in predictions introduced by the randomness, which may lead to the misclassification of the noise copies of test samples.
Additionally, the robustness space of any dynamic model (orange bars) is the intersection of the robustness spaces under all statuses of the dynamic model. 
Consequently, the dynamic model has a smaller robustness space compared to the original model.
Among dynamic models, the dynamic attention model demonstrates the highest robustness across various perturbation radii, thereby effectively preserving its robustness.
The introduction of dropout, however, decreases the robustness of the original model, which suggests that dropout introduces excessive randomness, leading to unstable predictions and an increased likelihood of misclassifying noise samples, as further supported by Sec. \ref{sec:stable}.}

Interestingly, the fusion model (red bars) improves the robustness of the dropout model (green bars) in most cases. 
We speculate that removing tokens with high attention values helps prevent certain neurons from containing excessive amounts of textual information while others do not. 
Consequently, while dropout disables some neurons, the remaining valid neurons contain enough textual information to facilitate information flow to subsequent layers.
\SLJ{As the modification rates (shown in Fig. \ref{fig:robustness_0.2} and \ref{fig:robustness_0.4} in Appendix \ref{appendix:robustness}) or perturbations increase, the dynamic attention model consistently maintains a similar level of robustness to the original model. }
In contrast, the robustness of the dropout and fusion models diminishes rapidly.

Overall, the dynamic attention model can preserve 98\%, on average, of the original model's robustness and retain the most robustness among all three dynamic models. 
In contrast, the dropout and fusion models can only preserve 83\%, on average, of the original robustness.

%-------------------------------------------------------------------------------
\section{Challenges and Future Works}\label{sec:discussion}
%-------------------------------------------------------------------------------
%\subsection{Challenges}\label{sec:challenges}
%Despite the success of dynamic modeling techniques, only a few researches have been studied on them from the theoretical perspective.
%Our dynamic attention shares the same challenge.
\SLJ{Dynamic modeling techniques, while successful, are not extensively explored theoretically, and our dynamic attention faces similar challenges. 
For instance, our robustness analysis can only be evaluated under relaxed conditions. 
In addition, the attention map's masking introduces discrepancies between the training and inference since no tokens are masked during training, which can reduce the original task's performance.}
%This may lead to a decrease in the performance of the original task, and previous experiments have also shown that the accuracy of the dynamic attention model is slightly lower than that of the original model.

%\subsection{Future Works}\label{sec:future}
In this paper, dynamic attention is designed to mitigate the effect of adversarial attacks by masking attentive tokens in the attention map. 
However, these attentive tokens are also likely to be critical in normal texts. 
Therefore, if tokens can be selected more precisely, the impact of adversarial attacks can be further reduced, and the original task performance will not be affected.
%Additionally, in the masking process, we directly mask all attention values of the selected key tokens. 
%Future studies might examine the effect of each attention value on the hidden representations and the output confidences and make precise adjustments.
%Moreover, there is a discrepancy between training and testing, as stated in Sec. \ref{sec:challenges}, which can be alleviated by employing dynamic attention during the pre-training or fine-tuning phase.
Additionally, the performance of dynamic attention can also be further improved by integrating with other robustness-enhancement modules and utilizing other defense techniques such as detection and restoration \cite{mozes2021frequency, wang2022detecting, zhou2019learning, li2020textshield} to preprocess the input before fed into the model. 
%further reduce the attack success rate.
%Other types of defenses like adversarial examples detection and restoration methods \cite{mozes2021frequency, wang2022detecting, zhou2019learning, li2020textshield} can also be implemented before the input text is fed into the target model to reduce the attack success rate.

%-------------------------------------------------------------------------------
\section{Conclusion}
\label{sec:conclusion}
%-------------------------------------------------------------------------------
In this work, we propose dynamic attention, the first dynamic modeling method tailored for transformer-based models. 
Our method reduces the impact of adversarial examples by dynamically masking or weakening highly attentive tokens in the input text. 
Additionally, it can be combined with the dropout module, information bottleneck, adversarial training, and other modules to further reduce the attack success rate and increase the attack difficulty. 
Our experiments on three classification and two generation tasks demonstrate that dynamic attention is effective in defending against adversarial attacks under various scenarios, even with no attacker knowledge. 
We also demonstrate its capability to defend against adversarial attacks under data shift and adaptive attack scenarios.
Furthermore, dynamic attention is effective in attenuating the impact of backdoor triggers.
%We also consider three different attack scenarios depending on the attacker's knowledge of the target model and study the performance under these scenarios.
Finally, our results reveal that dynamic attention maintains stability in repeated predictions and preserves the robustness of the original model.


\section*{Acknowledgment}
We thank the anonymous reviewers for their valuable comments to improve our paper.
This work was partly supported by the National Key Research and Development Program of China under No. 2022YFB3102100, NSFC under No. 62102360, 62076125, U20B2049 and 62102360, and Open Research Projects of Zhejiang Lab (No. 2022RC0AB01).



%\bibliographystyle{IEEEtranS}
%\begin{thebibliography}{1}
%\end{thebibliography}
\bibliographystyle{plain}
\bibliography{bib.bib}
\newpage
\appendix

\subsection{The Performance of Text Generation with Dynamic Modeling Enabled in Both Encoder and Decoder}\label{appendix:encoder-decoder}
We primarily chose to implement attention masking solely on the encoder side based on preliminary observations that enabling dynamic modeling in the T5's decoder will degrade the model's performance.
To substantiate our approach, we conducted comparative experiments by enabling dynamic modeling in both the encoder and the decoder. 
The results, which can be found in Table \ref{table:encoder-decoder}, clearly show a significant degradation in performance on adversarial texts when dynamic modeling is enabled on both ends. 
%This is mainly because the masking or weakening of tokens in the early stage of generation will significantly affect the following generation. 
%Thus, we disabled dynamic modeling within the decoder in the first manuscript. 

\begin{figure}[] 
    \centering
    \includegraphics[scale=0.24]{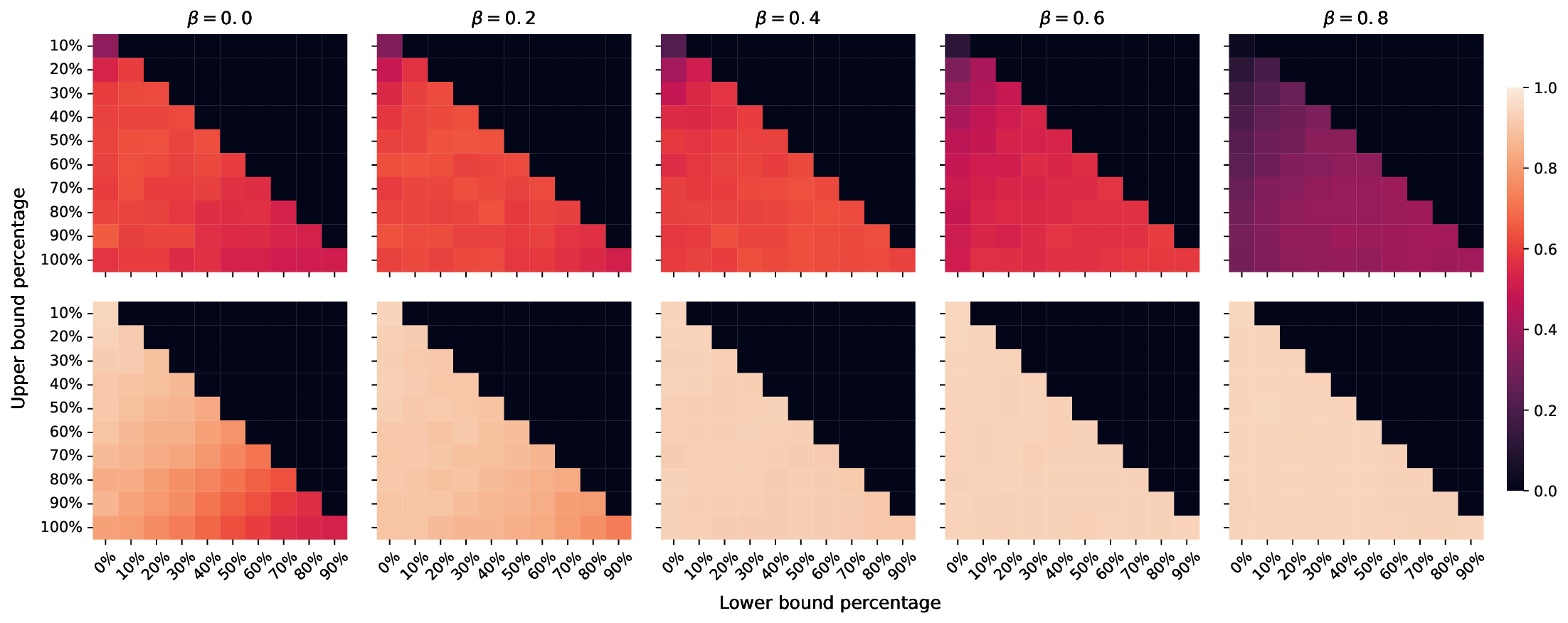}
    \caption{\SLJ{The prediction accuracy of adversarial texts (upper row) and the prediction accuracy of clean texts (bottom row) from the Amazon dataset under the {\em prefix-tuned} model with different $m$'s range and different $\beta$ value.}}
    \label{fig:amazon_m_prefix}
    \includegraphics[scale=0.24]{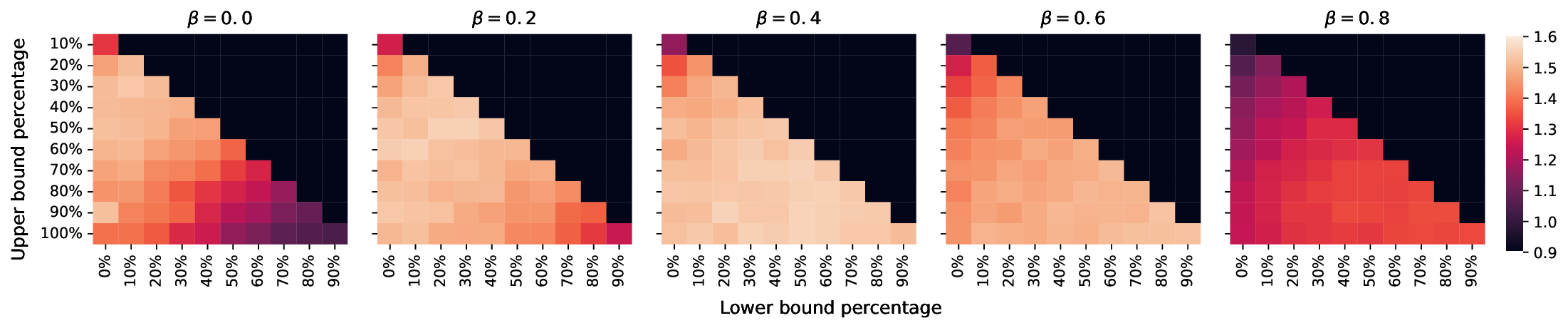}
    \caption{\SLJ{The overall metric $M$ of the dynamic attention model trained with the Amazon dataset under the {\em prefix-tuned} model with different $m$'s range and different $\beta$ value.}}
    \label{fig:amazon_b_prefix}
\end{figure}

\begin{figure}[] 
    \centering
    \includegraphics[scale=0.24]{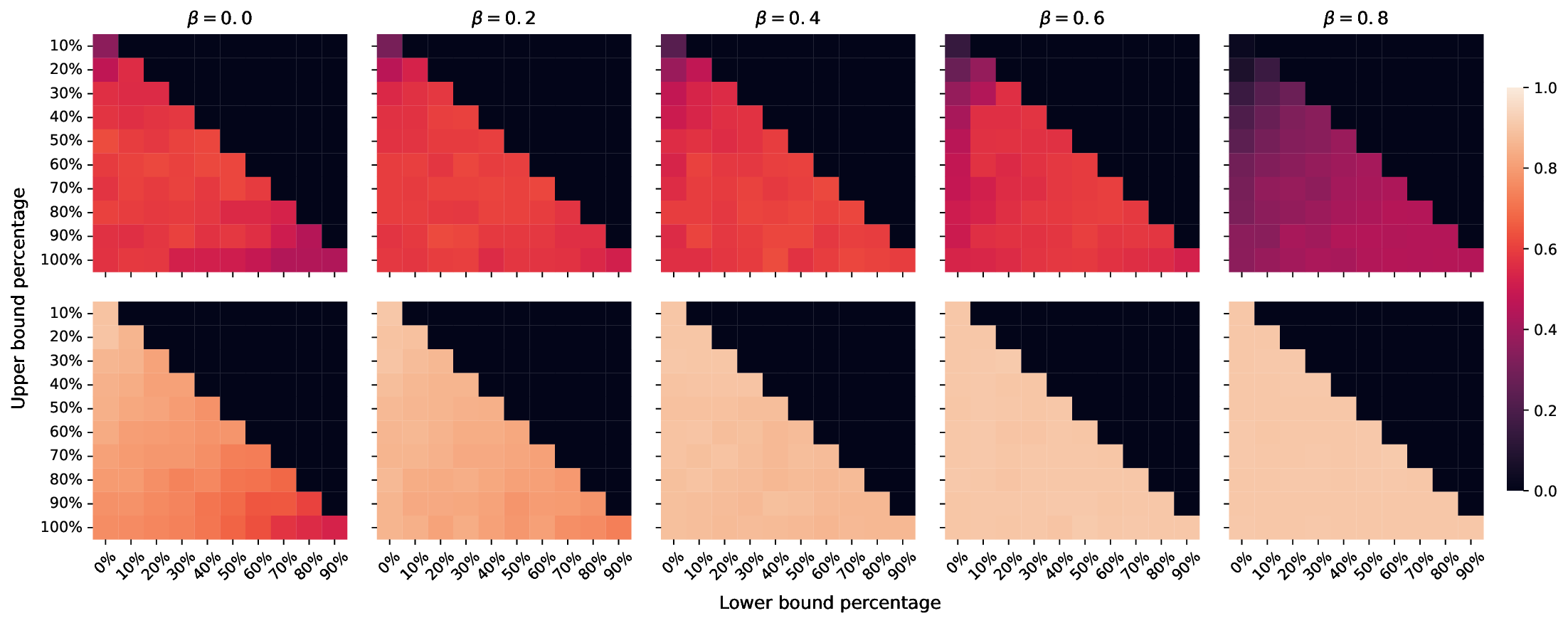}
    \caption{\SLJ{The prediction accuracy of adversarial texts (upper row) and the prediction accuracy of clean texts (bottom row) from the Amazon dataset under the {\em prompt-tuned} model with different $m$'s range and different $\beta$ value.}}
    \label{fig:amazon_m_prompt}
    \includegraphics[scale=0.24]{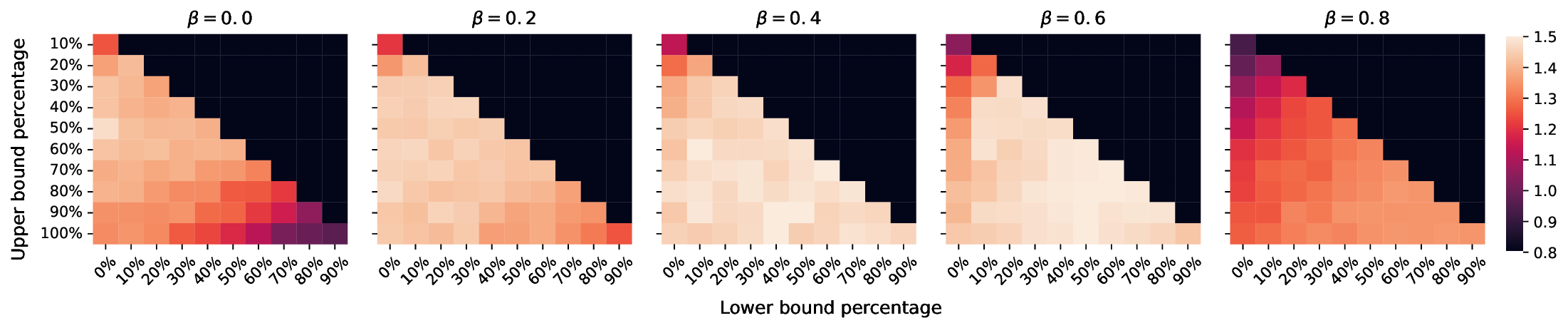}
    \caption{\SLJ{The overall metric $M$ of the dynamic attention model trained with the Amazon dataset under the {\em prompt-tuned} model with different $m$'s range and different $\beta$ value.}}
    \label{fig:amazon_b_prompt}
\end{figure}

\subsection{Sensitivity Analysis on Other Models and Tasks}\label{appendix:sensitivity}
\begin{figure}[] 
    \centering
    \includegraphics[scale=0.24]{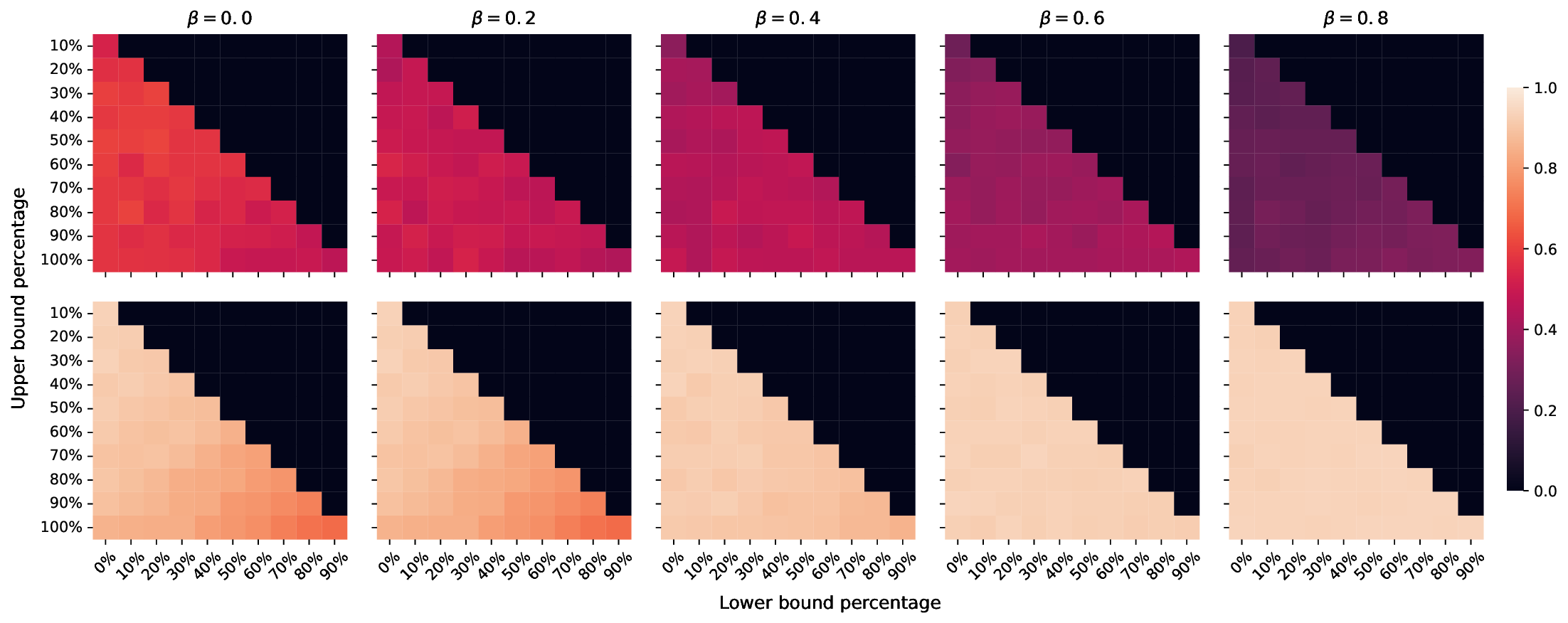}
    \caption{\SLJ{The prediction accuracy of adversarial texts (upper row) and the prediction accuracy of clean texts (bottom row) from the Amazon dataset under the GPT-2 model with different $m$'s range and different $\beta$ value.}}
    \label{fig:gpt_m}
    \vspace{-5mm}
\end{figure}

\begin{figure}[] 
    \centering
    \includegraphics[scale=0.24]{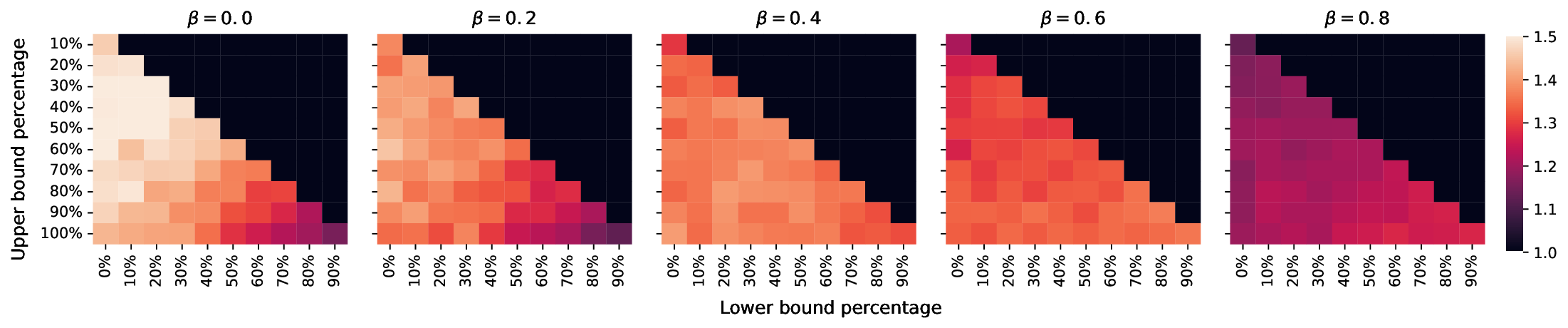}
    \caption{\SLJ{The overall metric $M$ of the dynamic attention model trained with the Amazon dataset under the GPT-2 model with different $m$'s range and different $\beta$ value.}}
    \label{fig:gpt_b}
\end{figure}

For text classification, similar to the fine-tuned Amazon model, we have conducted sensitivity analysis on the prefix-tuned model and prompt-tuned model. 
The results of the prefix-tuned model and prompt-tuned model are shown in Figs. \ref{fig:amazon_m_prefix}, \ref{fig:amazon_b_prefix}, \ref{fig:amazon_m_prompt}, \ref{fig:amazon_b_prompt}
, \ref{fig:gpt_m}, and \ref{fig:gpt_b}. 
%We have also extended our scrutiny to another dataset, Twitter, with results shown in Figs. \ref{fig:twitter_m} and \ref{fig:twitter_b}.}

Similar observations can be found for the prefix-tuned model and the fine-tuned Twitter model.
For the prompt-tuned model, it is appropriate to use a range of 20\% to 40\% for $m$ and set $\beta=0.4$.

\begin{figure}[] 
    \centering
    \includegraphics[scale=0.4]{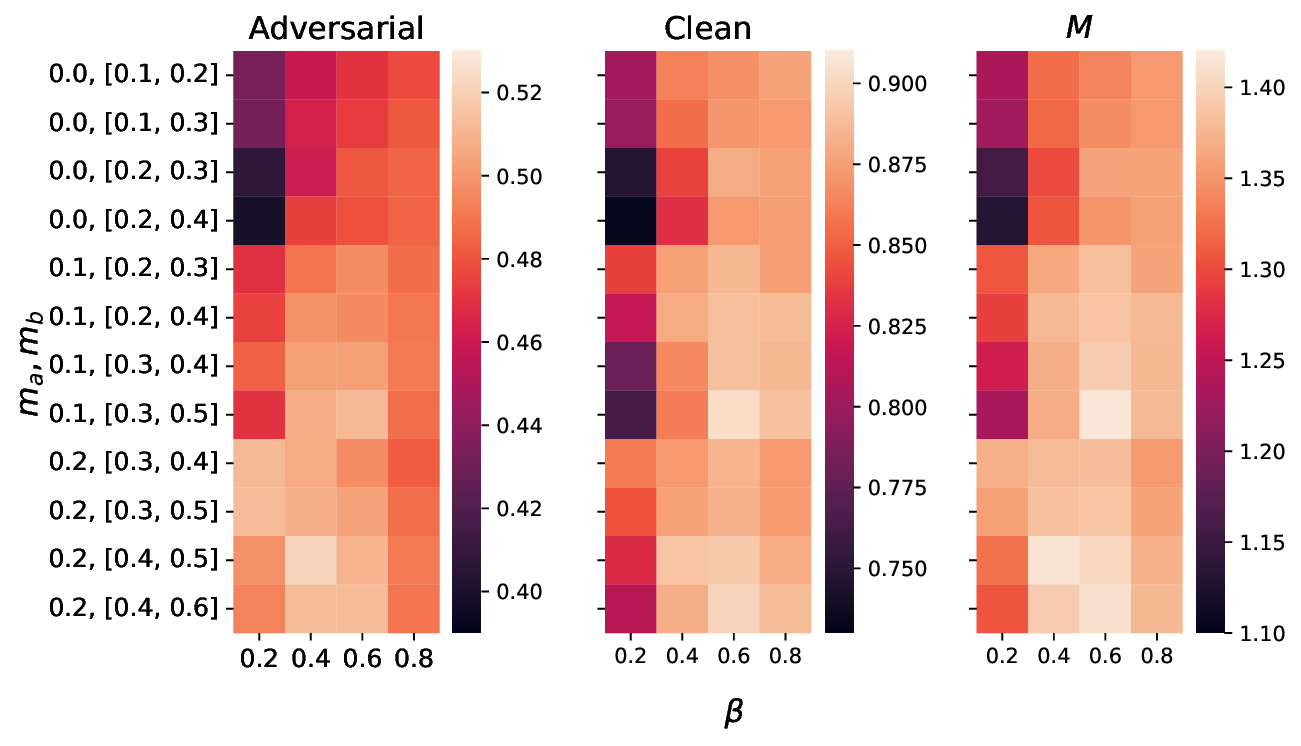}
    \caption{\SLJ{The overall metric $M$ of the dynamic attention model trained with the Twitter dataset under different $m$'s range and different $\beta$ value (the higher, the better).}}\label{fig:text_generation_adv}
\end{figure}

Moreover, for text generation, we introduce a sensitivity analysis concerning the range of random variable $m_b$ and the selection of parameters $m_a$ and $\beta$, where tokens rank between $m_a$ and $m_b$ are chosen to be masked or weakened in text generation scenarios. 
Similar to the choice of text classification scenario, we determine $m_a$ and $m_b$ to be a proportion of input text length.
We vary the $\beta$ from 0.2 to 0.8 with an increment of 0.2 and $m_a$ from 0.0 to 0.2 with an increment of 0.1. 
For the range of $m_b$, we pre-defined four ranges for each $m_a$.
We use adversarial examples from the English to French dataset for illustration, and the BLEU scores for each hyperparameter setting are shown in Fig. \ref{fig:text_generation_adv}.

From the figure, it can be observed that choosing $m_a=0.1$, $\beta=0.6$, and setting the range of $m_b$ to be $[0.3, 0.5]$ achieves the best performance in text generation and outperforms the translation using the original static model.
This result is consistent with our previous choice of keeping the top few tokens unchanged and masking or weakening later tokens.

\subsection{The Performance of Dynamic Attention under Varying Confidence}\label{appendix:varying_confidence}
To underscore the effectiveness of dynamic attention mechanisms, we conduct experiments on adversarial examples with varying confidence levels, probing the dynamic model's ability to discern and respond to adversarial perturbations. 
We collect adversarial texts generated by TextBugger, TextFooler and PWWS from the fine-tuned model under the Amazon dataset and separate them into eight categories according to their confidence level, which are: {\em (1).} $\text{confidence}<0.65$, {\em (2).} $0.65<\text{confidence}<0.70$, {\em (3).} $0.70<\text{confidence}<0.75$, {\em (4).} $0.75<\text{confidence}<0.80$, {\em (5).} $0.80<\text{confidence}<0.85$, {\em (6).} $0.85<\text{confidence}<0.90$, {\em (7).} $0.90<\text{confidence}<0.95$, and {\em (8).} $0.95<\text{confidence}<1.00$. 
The ASR of adversarial texts for each category under the dynamic attention model and dropout model is presented in Table \ref{table:varing_confidence}.
Note that the ASR of these adversarial texts is 100\% in the static model.

The Table shows that both the dynamic attention model and dropout model perform better on adversarial texts with low confidence values, leading to a lower ASR. 
However, the dynamic attention model consistently outperforms the dropout model, particularly evident at high confidence levels. (i.e. $0.95<\text{confidence}<1.00$).

\begin{table*}[t]
\centering
\caption{\SLJ{The BLEU score of adversarial texts generated from local static T5 model using TED Talk and Gigawords. The `Encoder' row indicates the dynamic modeling is enabled only in the encoder of the T5 model. The `Encoder \& Decoder' row indicates the dynamic modeling is enabled in both encoder and decoder of the T5 model.}}
\SLJ{\begin{tabular}{cccccc}
\toprule
\multirow{2}{*}{Task}              & \multirow{2}{*}{Model type} & \multicolumn{2}{c}{TextBugger} & \multicolumn{2}{c}{TextFooler} \\
\cline{3-6}
&                             & Encoder  & Encoder \& Decoder  & Encoder  & Encoder \& Decoder  \\
                                   \hline
\multirow{4}{*}{English to French} & original model              & 0.4698   & 0.4698              & 0.4981   & 0.4981              \\
& dynamic attention           & 0.5034   & 0.4641              & 0.5262   & 0.4661              \\
& dropout                     & 0.3701   & 0.3411              & 0.3967   & 0.3255              \\
& fusion model                & 0.3983   & 0.3195              & 0.3899   & 0.3131              \\
                                   \hline
\multirow{4}{*}{English to German} & original model              & 0.3810   & 0.3810              & 0.3715   & 0.3715              \\
& dynamic attention           & 0.4209   & 0.4052              & 0.4258   & 0.2760              \\
& dropout                     & 0.3179   & 0.2525              & 0.3291   & 0.2387              \\
& fusion model                & 0.3472   & 0.2614              & 0.3592   & 0.2378              \\
                                   \hline
\multirow{4}{*}{Summarization}     & original model              & 0.6197   & 0.6197              & 0.6307   & 0.6307              \\
& dynamic attention           & 0.6463   & 0.5806              & 0.6511   & 0.5575              \\
& dropout                     & 0.5347   & 0.4129              & 0.5298   & 0.4196              \\
& fusion model                & 0.5081   & 0.4199              & 0.5137   & 0.4026     \\
                                   \bottomrule        
\end{tabular}}\label{table:encoder-decoder}
\end{table*}

\begin{table*}[]
\centering
\caption{\SLJ{The performance of the dynamic attention model and the dropout model under varying confidence.}}
\scalebox{1}{
\SLJ{\begin{tabular}{ccccccc}
\toprule                      
& \multicolumn{2}{c}{Fine-tuning} & \multicolumn{2}{c}{Prefix-tuning} & \multicolumn{2}{c}{Prompt-tuning} \\
\cline{2-7}
& dynamic attetion    & dropout   & dynamic attetion     & dropout    & dynamic attetion     & dropout    \\
\hline
$\text{confidence}<0.65$       & 13.11\%             & 31.15\%   & 34.19\%              & 45.24\%    & 36.94\%              & 41.89\%    \\
$0.65<\text{confidence}<0.70$  & 19.15\%             & 38.30\%   & 40.30\%              & 46.77\%    & 38.24\%              & 46.08\%    \\
$0.70<\text{confidence}<0.75$  & 20.93\%             & 25.58\%   & 42.06\%              & 57.14\%    & 50.82\%              & 52.46\%    \\
$0.75<\text{confidence}<0.80$  & 37.50\%             & 46.88\%   & 41.79\%              & 55.22\%    & 51.95\%              & 54.55\%    \\
$0.80<\text{confidence}<0.85$  & 18.92\%             & 35.14\%   & 47.73\%              & 75.00\%    & 68.97\%              & 72.41\%    \\
$0.85<\text{confidence}<0.90$  & 29.55\%             & 38.64\%   & 55.68\%              & 71.59\%    & 72.73\%              & 72.73\%    \\
$0.90<\text{confidence}<0.95$  & 24.53\%             & 39.62\%   & 58.00\%              & 88.00\%    & 80.77\%              & 71.15\%    \\
$0.95<\text{confidence}<1.00$  & 42.62\%             & 68.31\%   & 67.01\%              & 88.66\%    & 77.42\%              & 90.32\%       \\
\bottomrule
\end{tabular}}}\label{table:varing_confidence}
\end{table*}

%\begin{figure}[] 
%    \centering
%    \includegraphics[scale=0.24]{figs/twitter_hyperparameter_m.eps}
%    \caption{\SLJ{The prediction accuracy of adversarial texts (upper row) and the prediction accuracy of clean texts (bottom row) from the Twitter dataset under different $m$'s range and different $\beta$ value (the higher, the better).}}
%    \label{fig:twitter_m}
%    \vspace{-5mm}
%\end{figure}
%
%
%\begin{figure}[] 
%    \centering
%    \includegraphics[scale=0.24]{figs/twitter_hyperparameter_b.eps}
%    \caption{\SLJ{The overall metric $M$ of the dynamic attention model trained with the Twitter dataset under different $m$'s range and different $\beta$ value (the higher, the better).}}
%    \label{fig:twitter_b}
%\end{figure}

\begin{figure}[]
	\centering
    \includegraphics[scale=0.65]{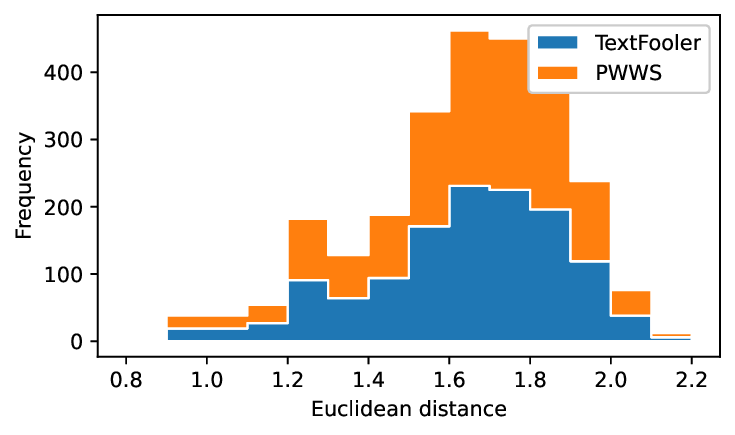}
    \caption{The stacked frequency histogram of Euclidean distance between original word and replaced word by TextFooler and PWWS attack.}
    \label{fig:dist}
\end{figure}

\begin{figure*}[]
	\centering
    \includegraphics[scale=0.65]{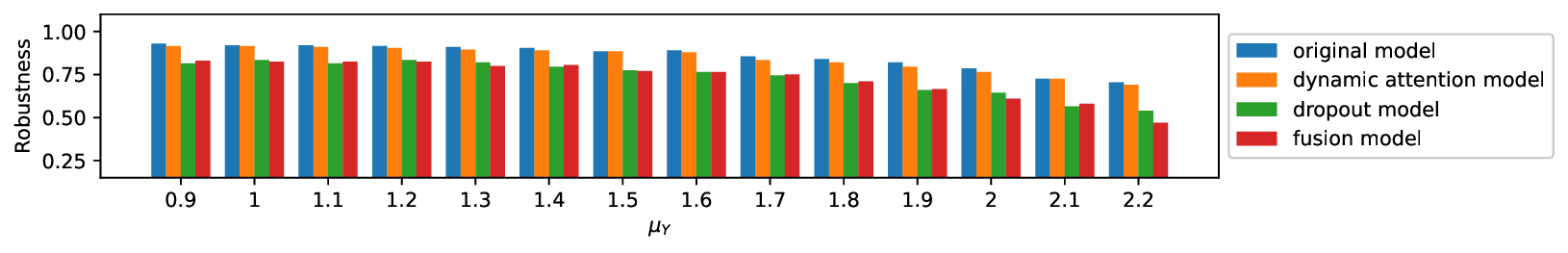}
    \caption{The percentage of robust samples under the attack of 20\% modification rates.}
    \label{fig:robustness_0.2}
    \includegraphics[scale=0.65]{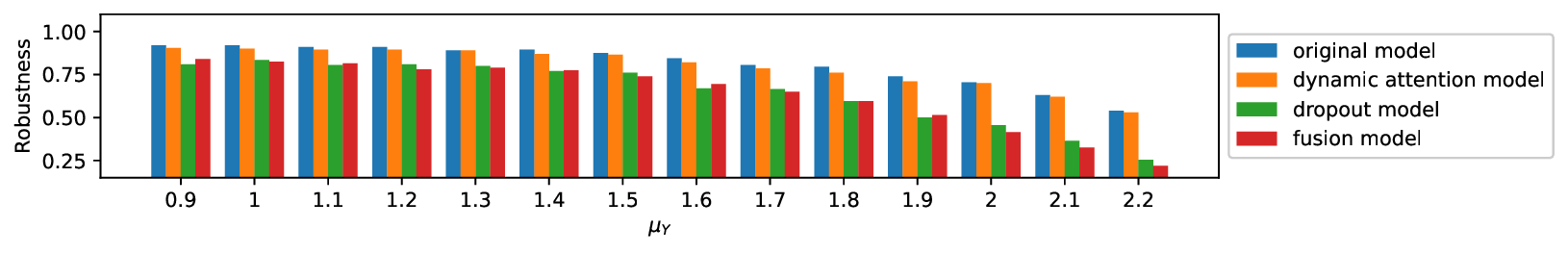}
    \caption{The percentage of robust samples under the attack of 40\% modification rates.}
    \label{fig:robustness_0.4}
    \vspace{-5mm}
\end{figure*}

\subsection{Euclidean Distance between the Original Words and Replaced Words}\label{appendix:euclidean}
In order to correspond to the actual attack scenario, we determine the range of $\sigma$ used in Sec. \ref{sec:analysis} according to the perturbation size of the actual adversarial attacks.
As representatives, we use synonym replacement attacks, such as TextFooler and PWWS.
The distribution of the Euclidean distance between the original words and replaced words under TextFooler and PWWS is shown in Fig. \ref{fig:dist}.
The figure shows that the distance between the original and replaced words typically falls within the range of 0.9 to 2.2. 

\subsection{Robustness Analysis under Different Modification Rates}\label{appendix:robustness}
The statistical robustness under 20\% and 40\% modification rates is provided in Figs. \ref{fig:robustness_0.2} and \ref{fig:robustness_0.4}, respectively.
From these figures, the dynamic attention model outperforms other dynamic models in robustness across different perturbation radii, emphasizing its efficacy in preserving robustness.
However, introducing dropout diminishes the original model's robustness, possibly due to its injection of randomness, leading to unstable predictions.
Notably, the fusion model (denoted by red bars) often enhances the dropout model's (green bars) robustness. 
These results are consistent with previous experimental results under 10\% modification rates.

\end{document}